\documentclass[10pt,journal,compsoc]{IEEEtran}

\ifCLASSOPTIONcompsoc
  \usepackage[nocompress]{cite}
\else
  \usepackage{cite}
\fi

\usepackage{floatrow}
\usepackage[font=footnotesize,skip=5pt]{caption}
\usepackage{epsfig}
\usepackage{graphicx}
\usepackage{amsmath}
\usepackage{amssymb}
\usepackage{algorithm}
\usepackage{algorithmic}
\usepackage{multirow}
\usepackage{color}
\usepackage{epstopdf}
\usepackage{booktabs}
\usepackage{pifont}

\usepackage{tabulary,overpic,xcolor}
\usepackage[caption=false]{subfig}
\definecolor{citecolor}{RGB}{34,139,34}
\usepackage[pagebackref=true,breaklinks=true,letterpaper=true,colorlinks,
  citecolor=citecolor,bookmarks=false]{hyperref}

\newcommand{\cmark}{\ding{51}}
\newcommand{\xmark}{\ding{55}}

\newcommand{\bd}[1]{\textbf{#1}}
\newcommand{\app}{\raise.17ex\hbox{$\scriptstyle\sim$}}

\newcolumntype{x}[1]{>{\centering\arraybackslash}p{#1pt}}

\newlength\savewidth\newcommand\shline{\noalign{\global\savewidth\arrayrulewidth
  \global\arrayrulewidth 1pt}\hline\noalign{\global\arrayrulewidth\savewidth}}
\newcommand{\tablestyle}[2]{\setlength{\tabcolsep}{#1}\renewcommand{\arraystretch}{#2}\centering\footnotesize}
\makeatletter\renewcommand\paragraph{\@startsection{paragraph}{4}{\z@}
  {.5em \@plus1ex \@minus.2ex}{-.5em}{\normalfont\normalsize\bfseries}}\makeatother

\begin{document}
%
\title{Cascade R-CNN: High Quality Object Detection and Instance
Segmentation}

\author{Zhaowei~Cai,
        and~Nuno~Vasconcelos
\IEEEcompsocitemizethanks{\IEEEcompsocthanksitem Z. Cai and N. Vasconcelos are with the Department of Electrical and Computer Engineering, University of California, San Diego, San Diego, CA 92093, USA,
E-mail: \{zwcai,nuno\}@ucsd.edu.
}
}

%
%


\IEEEtitleabstractindextext{%
\begin{abstract}
In object detection, the intersection over union (IoU) threshold is frequently used to define positives/negatives. The threshold used to train a detector defines its \textit{quality}. While the commonly used threshold of 0.5 leads to noisy (low-quality) detections, detection performance frequently degrades for larger thresholds. This paradox of high-quality detection has two causes: 1) overfitting, due to vanishing positive samples for large thresholds, and 2) inference-time quality mismatch between detector and test hypotheses. A multi-stage object detection architecture, the Cascade R-CNN, composed of a sequence of detectors trained with increasing IoU thresholds, is proposed to address these problems. The detectors are trained sequentially, using the output of a detector as training set for the next. This resampling progressively improves hypotheses quality, guaranteeing a positive training set of equivalent size for all detectors and minimizing overfitting. The same cascade is applied at inference, to eliminate quality mismatches between hypotheses and detectors. An implementation of the Cascade R-CNN without bells or whistles achieves state-of-the-art performance on the COCO dataset, and significantly improves high-quality detection on generic and specific object detection datasets, including VOC, KITTI, CityPerson, and WiderFace. Finally, the Cascade R-CNN is generalized to instance segmentation, with nontrivial improvements over the Mask R-CNN. To facilitate future research, two implementations are made available at \url{https://github.com/zhaoweicai/cascade-rcnn} (Caffe) and \url{https://github.com/zhaoweicai/Detectron-Cascade-RCNN} (Detectron).
\end{abstract}

}

\maketitle

\IEEEdisplaynontitleabstractindextext

%
\IEEEpeerreviewmaketitle

\section{Introduction}
\label{sec:intro}

Object detection is a complex problem, requiring the solution of two tasks.
First, the detector must solve the {\it recognition\/} problem, distinguishing
foreground objects from background and assigning them the proper
object class labels. Second, the detector must solve the
{\it localization\/} problem, assigning accurate bounding boxes to different
objects. An effective architecture for the solution of the two tasks,
on which many of the recently proposed object detectors are based, is the
two-stage R-CNN framework \cite{DBLP:conf/cvpr/GirshickDDM14,DBLP:conf/iccv/Girshick15,DBLP:conf/nips/RenHGS15,lin2017feature}.
This frames detection as a multi-task learning problem
that combines classification, to solve the recognition problem,
and bounding box regression, to solve localization.

\begin{figure}[!t]
\begin{minipage}[b]{.495\linewidth}
\centering
\centerline{\epsfig{figure=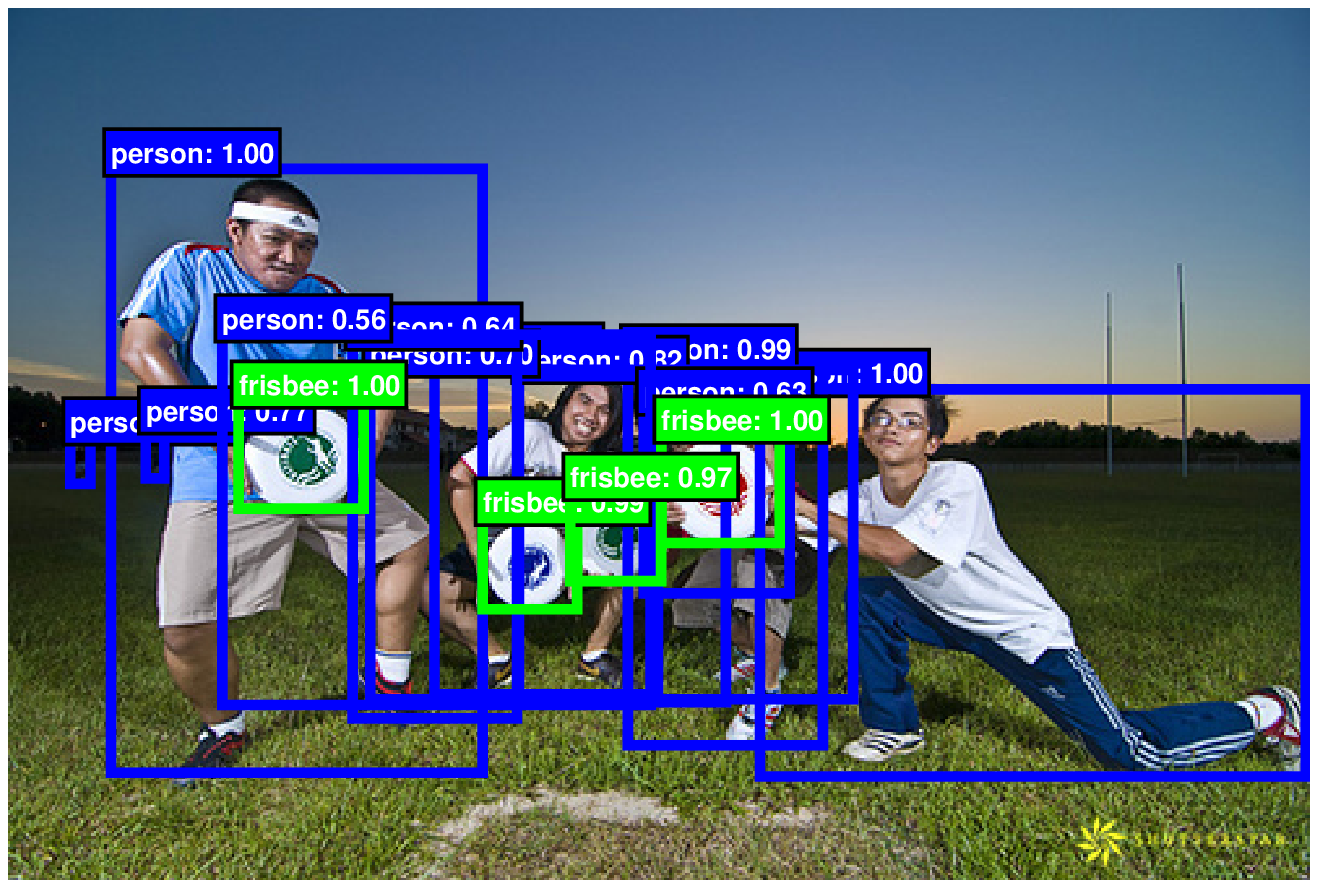,width=4.1cm,height=2.75cm}}{(a) Detection of $u=0.5$}
\end{minipage}
\hfill
\begin{minipage}[b]{.495\linewidth}
\centering
\centerline{\epsfig{figure=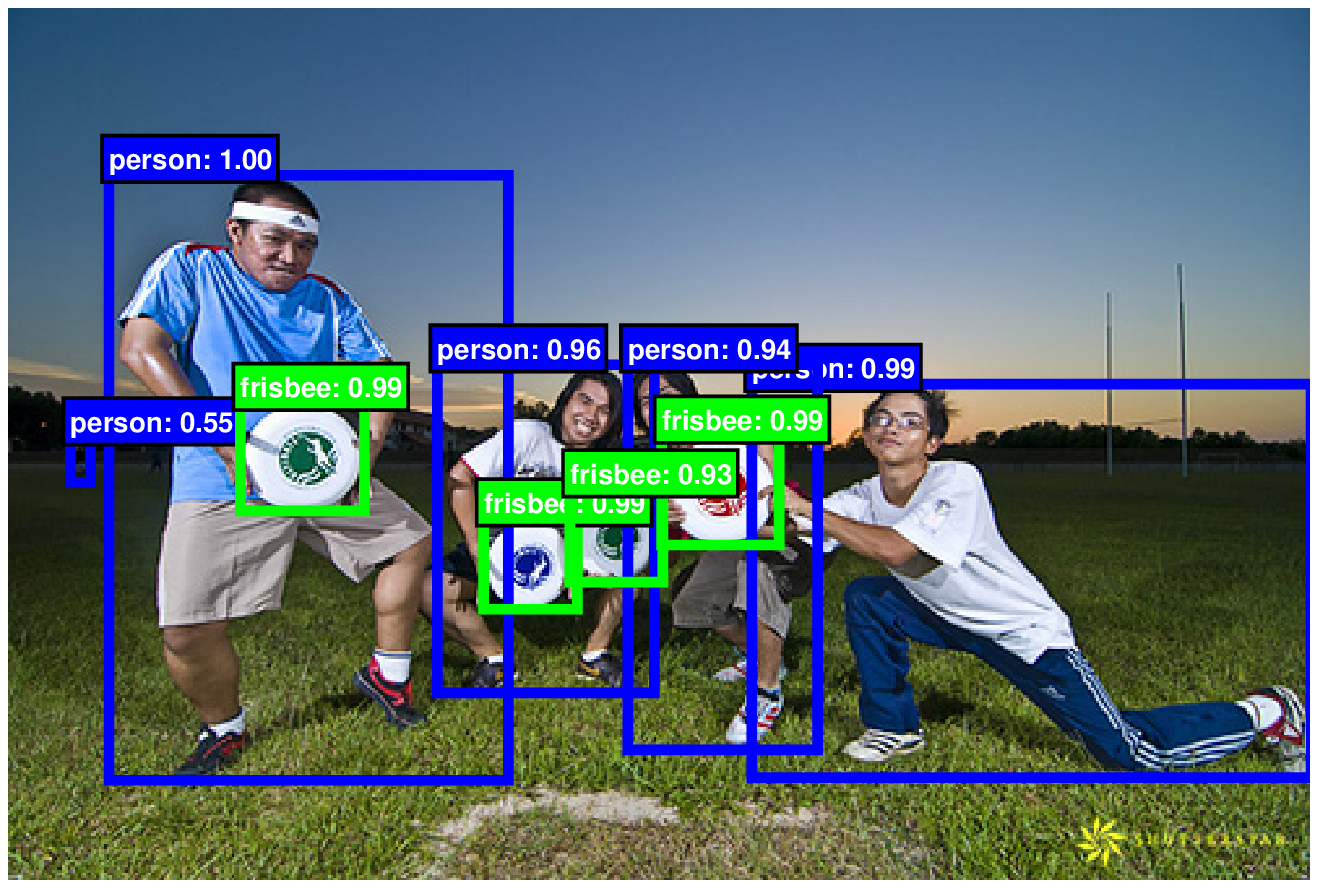,width=4.1cm,height=2.75cm}}{(b) Detection of $u=0.7$}
\end{minipage}\\\vspace{2mm}
\begin{minipage}[b]{.99\linewidth}
\centering
\centerline{\epsfig{figure=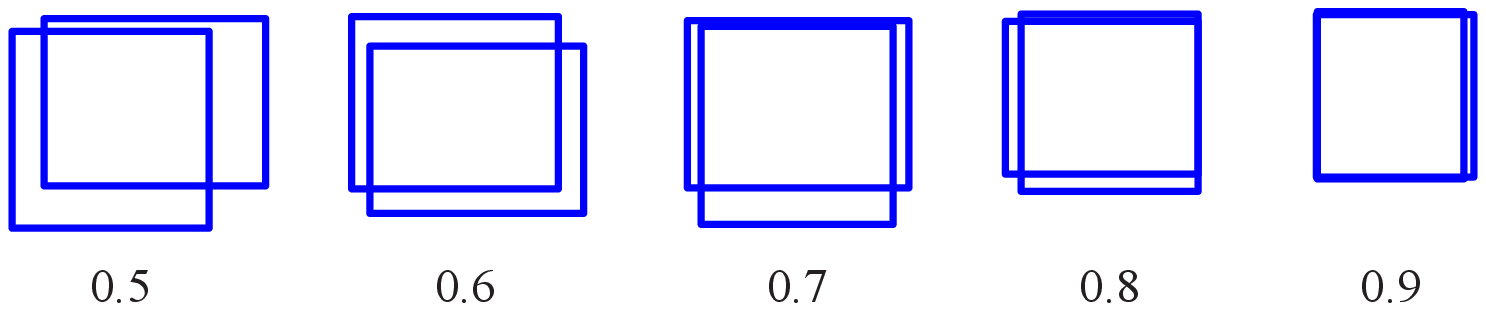,width=8cm,height=1.72cm}}{(c) Examples of increasing qualities}
\end{minipage}
\caption{(a) and (b) detections by object detectors of increasing qualities, and (c) examples of increasing quality.}
\label{fig:quality detection}
\end{figure}

Despite the success of this architecture, the two problems can be
difficult to solve accurately. This is partly due to the fact that there are
many ``close'' false positives, corresponding to ``close but not correct''
bounding boxes. An effective detector must find all true positives in an
image, while suppressing these close false positives. This requirement makes
detection more difficult than other classification problems, e.g. object
recognition, where the difference between positives and negatives is not as
fine-grained. In fact, the boundary between positives and negatives must
be carefully defined. In the literature, this is done
by thresholding the intersection over union (IoU) score between candidate
and ground truth bounding boxes. While the threshold is typically set at
the value of $u=0.5$, this is a very loose requirement for positives.
The resulting detectors frequently produce noisy bounding
boxes, as shown in Fig. \ref{fig:quality detection} (a).
Hypotheses that most humans would consider close false positives frequently pass
the $IoU\geq{0.5}$ test. While training examples assembled under the $u=0.5$ criterion are rich and diverse, they make it difficult to train detectors that can effectively
reject close false positives.

\begin{figure*}[!t]
\begin{minipage}[b]{.33\linewidth}
\centering
\centerline{\epsfig{figure=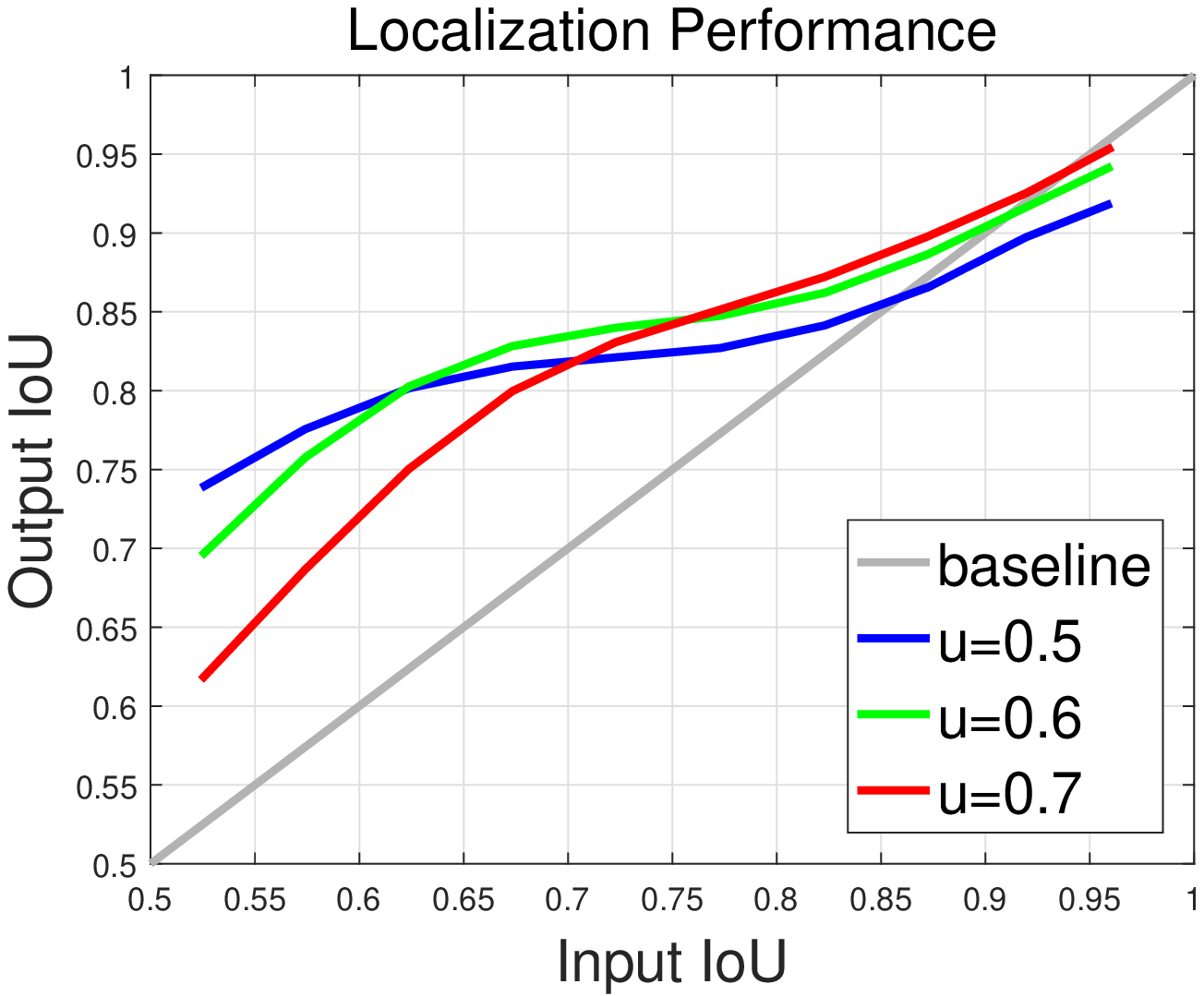,width=6cm,height=4.5cm}}{(a) Regressor}
\end{minipage}
\hfill
\begin{minipage}[b]{.33\linewidth}
\centering
\centerline{\epsfig{figure=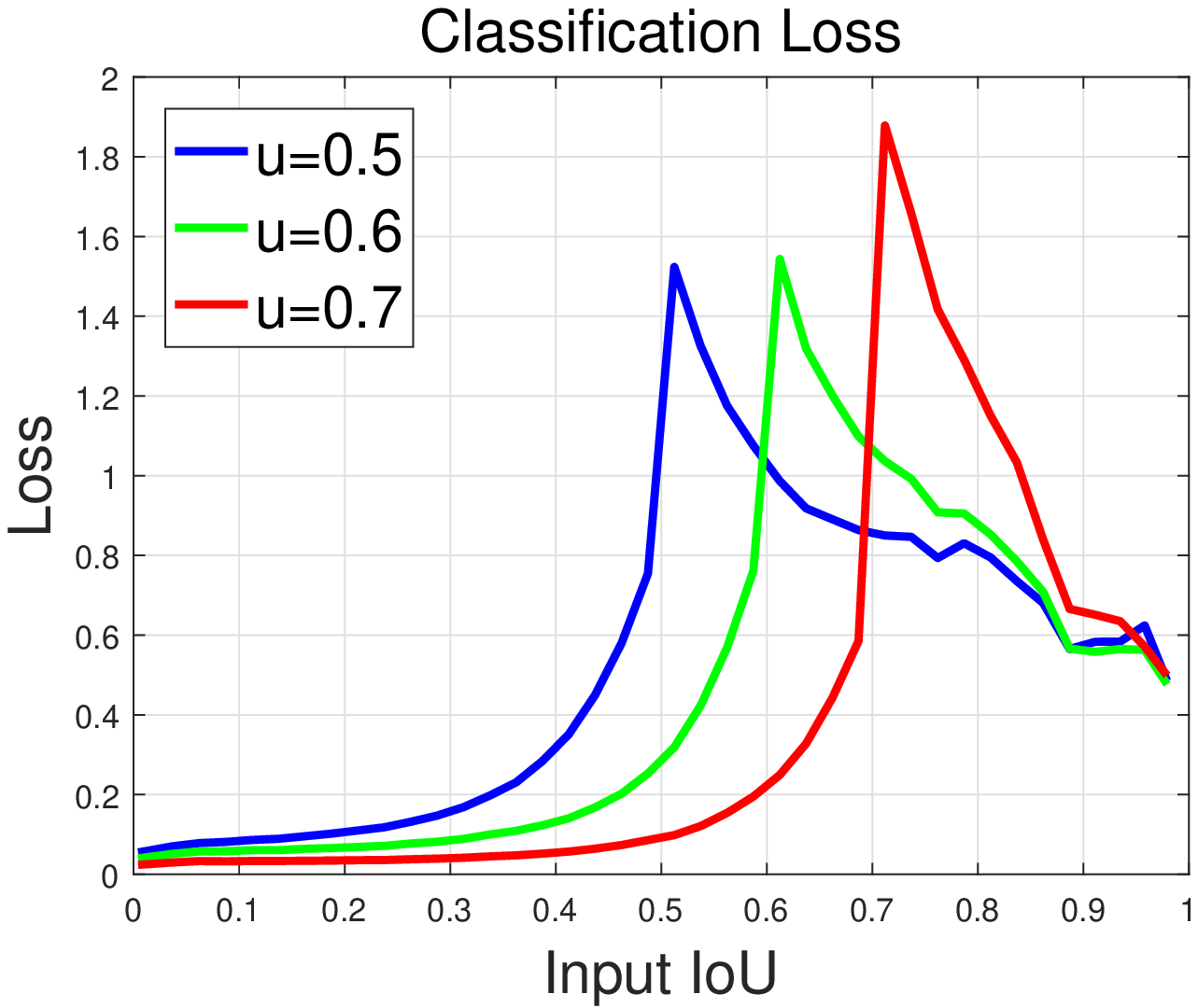,width=6cm,height=4.5cm}}{(b) Classifier}
\end{minipage}
\hfill
\begin{minipage}[b]{.33\linewidth}
\centering
\centerline{\epsfig{figure=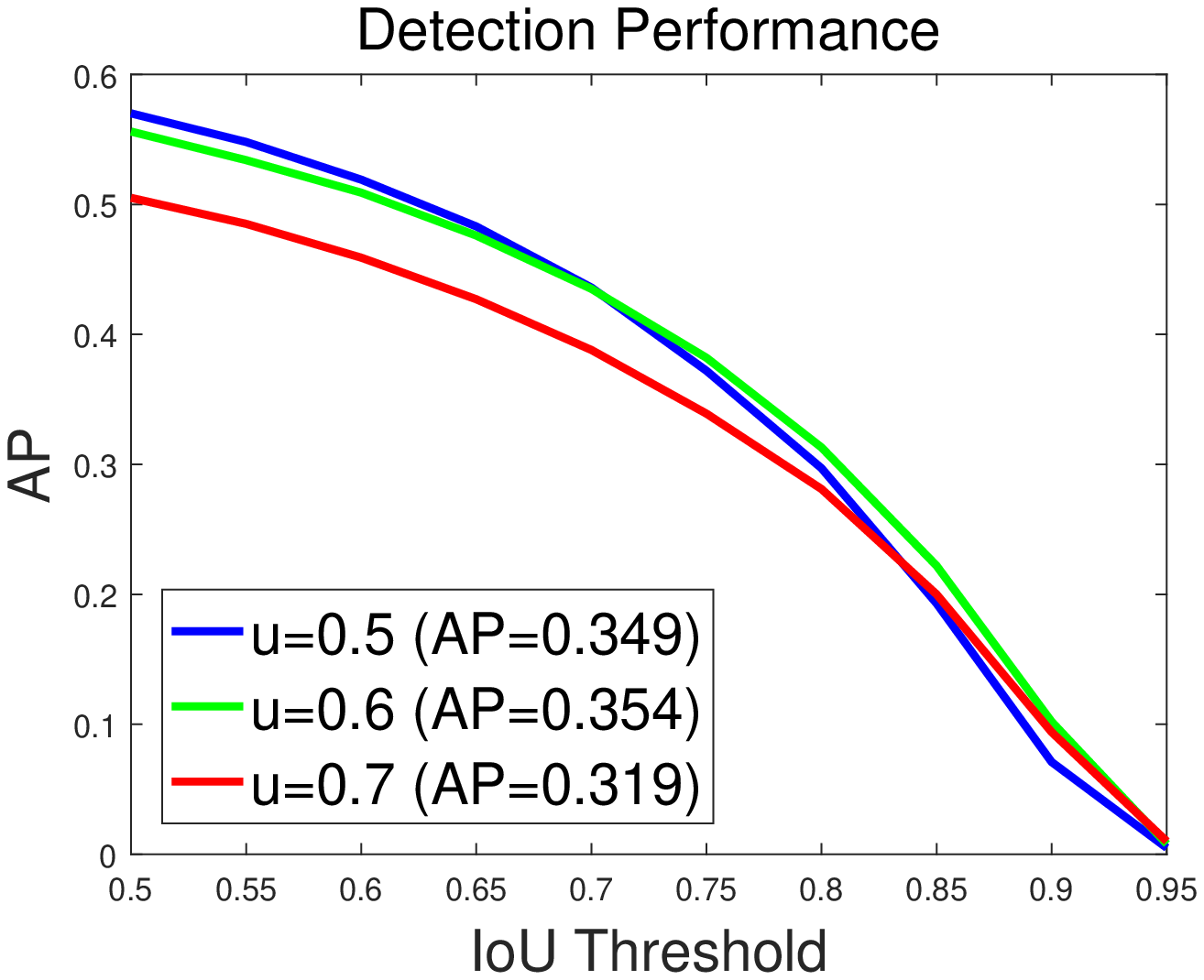,width=6cm,height=4.5cm}}{(c) Detector}
\end{minipage}
\caption{Bounding box localization, classification loss and detection performance of object detectors of increasing IoU threshold $u$.}
\label{fig:motivation}
\end{figure*}

In this work, we define the {\it quality\/} of a detection hypothesis as its
IoU with the ground truth, and the {\it quality of a detector\/} as the IoU
threshold $u$ used to train it. Some examples of hypotheses of increasing
quality are shown in  Fig. \ref{fig:quality detection} (c). The goal
is to investigate the poorly researched problem of learning
{\it high quality object detectors\/}. As shown in Fig.
\ref{fig:quality detection} (b), these are detectors that produce few close
false positives.
The starting premise is that a single detector can only be optimal for a
single quality level. This is known in the
cost-sensitive learning literature~\cite{DBLP:conf/ijcai/Elkan01,
DBLP:journals/pami/Masnadi-ShiraziV11}, where the optimization of
different points of the receiver operating characteristic (ROC) requires
different loss functions. The main difference is that we consider
the optimization for a given IoU threshold, rather than false positive rate.

Some evidence in support of this premise is given in Fig.
\ref{fig:motivation}, which presents the bounding box localization
performance, classification loss and detection performance, respectively, of
three detectors trained with IoU thresholds of $u=0.5,0.6,0.7$. Localization
and classification are evaluated as a function of the detection hypothesis
IoU. Detection is evaluated as a function of
the IoU threshold, as in COCO \cite{DBLP:conf/eccv/LinMBHPRDZ14}.
Fig. \ref{fig:motivation} (a)
shows that the three bounding box regressors tend to achieve the best
performance for examples of IoU in the vicinity of the threshold used for
detector training. Fig. \ref{fig:motivation} (c) shows a similar effect
for detection, up to some overfitting for the highest thresholds.
The detector trained with $u=0.5$
outperforms the detector trained with $u=0.6$ for low IoUs,
underperforming it at higher IoUs. In general, a detector optimized for
a single IoU value is not optimal for other values. This is also confirmed by
the classification loss, shown in Fig. \ref{fig:motivation} (b), whose
peaks are near the thresholds used for
detector training. In general, the threshold determines the classification
boundary where the classifier is most discriminative, i.e. has largest
margin \cite{DBLP:journals/ml/CortesV95,DBLP:conf/eurocolt/FreundS95}.

The observations above suggest that high quality detection requires a
close {\it match\/} between the quality of the detector and that of the
detection hypotheses. The detector will only achieve high quality if
presented with high quality proposals. This, however, cannot be guaranteed
by simply increasing the threshold $u$ during training. On the contrary,
as seen for the detector of $u=0.7$ in Fig. \ref{fig:motivation} (c),
forcing a high value of $u$ usually degrades detection performance.
We refer to this problem, i.e. that training a detector with higher
threshold leads to poorer performance, as the {\it paradox of
high-quality detection\/}. This problem has two causes. First, object
proposal mechanisms tend to produce  hypotheses distributions
heavily imbalanced towards low quality. In result, the use of larger
IoU thresholds during training exponentially reduces the number of positive
training examples. This is particularly problematic for neural networks,
which are very example intensive, making the ``high $u$'' training strategy
very prone to overfitting. Second, there is a mismatch between the quality
of the detector and that of the hypotheses available
at inference time. Since, as shown in Fig. \ref{fig:motivation}, high
quality detectors are only optimal for high quality hypotheses, detection
performance can degrade substantially for hypotheses of lower quality.

In this paper, we propose a new detector architecture, denoted
as {\it Cascade R-CNN,\/} that addresses these problems, to enable high quality
object detection. The new architecture is a multi-stage extension of the
R-CNN, where detector stages deeper into the cascade are sequentially more
selective against close false positives. As is usual for classifier
cascades~\cite{DBLP:journals/ijcv/ViolaJ04,DBLP:journals/pami/SaberianV12}, the cascade of R-CNN stages is trained
sequentially, using the output of one stage to train the next. This
leverages the observation that the output IoU of a bounding box
regressor is almost always better than its input IoU, as can be
seen in Fig. \ref{fig:motivation} (a), where nearly all plots are above
the gray line. In result, the output of a detector trained with a certain
IoU threshold is a good hypothesis distribution to train the detector of the
next higher IoU threshold. This has some similarity to {\it boostrapping\/}
methods commonly used to assemble datasets for object detection \cite{DBLP:journals/ijcv/ViolaJ04,DBLP:journals/pami/FelzenszwalbGMR10}. The main difference is that the
resampling performed by the Cascade R-CNN does not aim to mine hard negatives.
Instead, by adjusting bounding boxes, each stage aims to find a good set of
close false positives for training the next stage. The main outcome of
this resampling is that the quality of the detection hypotheses increases
{\it gradually\/}, from one stage to the next. In result, the sequence
of detectors addresses the two problems underlying
the paradox of high-quality detection. First, because the resampling
operation guarantees the availability of a {\it large number\/} of examples for
the training of all detectors in the sequence, it is possible
to train detectors of high IoU without overfitting. Second,
the use of the same cascade procedure at inference time produces a set of
hypotheses of progressively higher quality, well {\it matched\/} to the
increasing quality of the detector stages. This enables
higher detection accuracies, as suggested by Fig. \ref{fig:motivation}.

The Cascade R-CNN is quite simple to implement and trained end-to-end.
Our results show that a vanilla implementation, without any bells and whistles,
surpasses almost all previous state-of-the-art \emph{single-model} detectors, on the challenging COCO detection
task \cite{DBLP:conf/eccv/LinMBHPRDZ14}, especially under the stricter
evaluation metrics. In addition, the Cascade R-CNN can be built with any
two-stage object detector based on the R-CNN framework. We have
observed consistent gains (of 2$\sim$4 points, and more under stricter localization metrics), at a marginal increase in
computation. This gain is independent of the strength of the baseline object
detectors, for all the models we have tested. We thus believe that this simple and effective detection
architecture can be of interest for many object detection research efforts.

A preliminary version of this manuscript was previously published
in \cite{cai18cascadercnn}. After the original publication, the
Cascade R-CNN has been successfully reproduced within many different
codebases, including the popular Detectron \cite{Detectron2018}, PyTorch\footnote{https://github.com/open-mmlab/mmdetection}, and TensorFlow\footnote{https://github.com/tensorpack/tensorpack}, showing consistent and reliable improvements independently of
implementation codebase.
In this expanded version, we have extended the Cascade R-CNN to instance
segmentation, by adding a mask head to the cascade, denoted as {\it Cascade Mask R-CNN}.
This is shown to
achieve non-trivial improvements over the popular Mask
R-CNN \cite{he2017mask}. A new and more extensive evaluation
is also presented, showing that the Cascade R-CNN is compatible with many
complementary enhancements proposed in the detection and instance segmentation
literatures, some of which were introduced after \cite{cai18cascadercnn}, e.g. GroupNorm \cite{DBLP:conf/eccv/WuH18}. Finally, we further present the results of a larger set of
experiments, performed on various popular generic/specific object detection
datasets, including PASCAL VOC \cite{DBLP:journals/ijcv/EveringhamGWWZ10},
KITTI \cite{DBLP:conf/cvpr/GeigerLU12},
CityPerson \cite{DBLP:conf/cvpr/ZhangBS17} and
WiderFace \cite{DBLP:conf/cvpr/YangLLT16}. These experiments demonstrate
that the paradox of high quality object detection applies to all
these tasks, and that the Cascade R-CNN enables more effective high quality detection
than previously available methods. Due to these properties, as well as its
generality and flexibility, the Cascade R-CNN has recently been adopted
by the winning teams of the COCO 2018 instance segmentation
challenge\footnote{http://cocodataset.org/\#detection-leaderboard},
the OpenImage 2018
challenge\footnote{https://storage.googleapis.com/openimages/web/challenge.html}, and the Wider Challenge 2018\footnote{http://wider-challenge.org/}.
To facilitate future research, we have released the code on two codebases,
Caffe \cite{DBLP:conf/mm/JiaSDKLGGD14} and Detectron \cite{Detectron2018}.

\section{Related Work}

Due to the success of  the R-CNN \cite{DBLP:conf/cvpr/GirshickDDM14}
detector, which combines a proposal detector and a
region-wise classifier, this two-stage architecture has become predominant
in the recent past. To reduce redundant CNN computations, the
SPP-Net \cite{DBLP:conf/eccv/HeZR014} and Fast
R-CNN \cite{DBLP:conf/iccv/Girshick15} introduced
the idea of region-wise feature extraction, enabling the sharing of
the bulk of feature computations by object instances. The Faster
R-CNN \cite{DBLP:conf/nips/RenHGS15} then achieved further speeds-up by
introducing a region proposal network (RPN), becoming the cornerstone of
modern object detection. Later, some works extended this detector to
address various problems of detail. For example, the
R-FCN \cite{DBLP:conf/nips/DaiLHS16}
proposed efficient region-wise full convolutions to avoid the heavy CNN
computations of the Faster R-CNN; and the Mask R-CNN \cite{he2017mask} added
a network head that computes object masks to support instance segmentation.
Some more recent works have focused on normalizing feature
statistics \cite{peng2018megdet,DBLP:conf/eccv/WuH18}, modeling
relations between instances \cite{hu2018relation}, non maximum
suppression (NMS) \cite{DBLP:conf/iccv/BodlaSCD17}, and
other aspects \cite{DBLP:conf/cvpr/ShrivastavaGG16,liu2018path}.

Scale invariance, an important requisite for effective object detection,
has also received substantial attention in the
literature \cite{DBLP:conf/eccv/CaiFFV16,lin2017feature,singh2018analysis}.
While natural images contain objects at various scales, the
fixed receptive field size of the filters implemented by the
RPN \cite{DBLP:conf/nips/RenHGS15} makes it prone to scale mismatches.
To overcome this, the MS-CNN \cite{DBLP:conf/eccv/CaiFFV16} introduced a
multi-scale object proposal network, by generating outputs at multiple layers.
This leverages the different receptive field sizes of the different layers
to produce a set of scale-specific proposal generators, which is then
combined into a strong multi-scale generator. Similarly, the
FPN \cite{lin2017feature} detects high-recall proposals at multiple output layers, with recourse to a scale-invariant feature representation by adding a top-down
connection across feature maps of different network depths. Both the
MS-CNN and FPN rely on a feature pyramid representation for multi-scale object detection. SNIP \cite{singh2018analysis}, on the other hand,
recently revisited image pyramid in modern object detection. It normalizes the gradients from different object scales during training, such that the whole detector is scale-specific. Scale-invariant detection is
achieved by using an image pyramid at inference.

One-stage object detection architectures have also become popular
for their computational efficiency. YOLO \cite{DBLP:conf/cvpr/RedmonDGF16}
outputs very sparse detection results and enables real-time
object detection, by forwarding the input image once through an efficient
backbone network. SSD \cite{DBLP:conf/eccv/LiuAESRFB16}
detects objects in a way similar to the RPN \cite{DBLP:conf/nips/RenHGS15},
but uses multiple feature maps at different resolutions to cover objects at
various scales. The main limitation of these detectors is that their
accuracy is typically below that of two-stage detectors.
The RetinaNet \cite{lin2017focal} detector was proposed to address the extreme
foreground-background class imbalance of dense object detection,
achieving results  comparable to two-stage detectors. Recently,
CornerNet \cite{DBLP:conf/eccv/LawD18} proposed to detect an object bounding
box as a pair of keypoints, abandoning the widely used concept of anchors
first introduced by the Faster R-CNN. This detector has achieved very good
performance with the help of some training and testing enhancements.
RefineDet \cite{zhang2018single} added an anchor
refinement module to the single-shot SSD \cite{DBLP:conf/eccv/LiuAESRFB16},
to improve localization accuracy. This is somewhat similar to the cascaded
localization implemented by the proposed Cascade R-CNN, but ignores the
problem of high-quality detection.

\begin{figure*}[!t]
\begin{minipage}[b]{.17\linewidth}
\centering
\centerline{\epsfig{figure=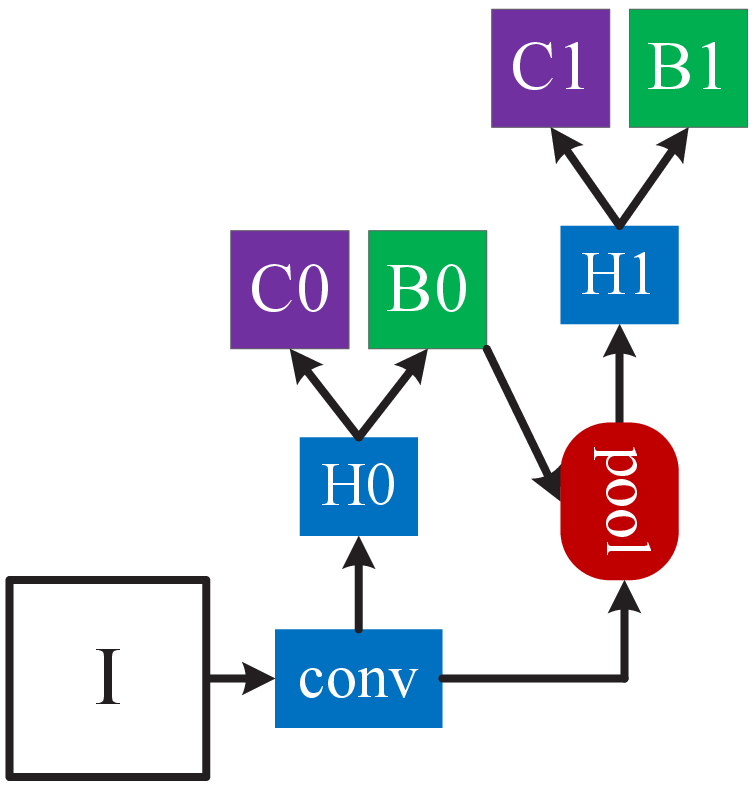,width=2.95cm,height=3.06cm}}{(a) Faster R-CNN}
\end{minipage}
\hfill
\begin{minipage}[b]{.3\linewidth}
\centering
\centerline{\epsfig{figure=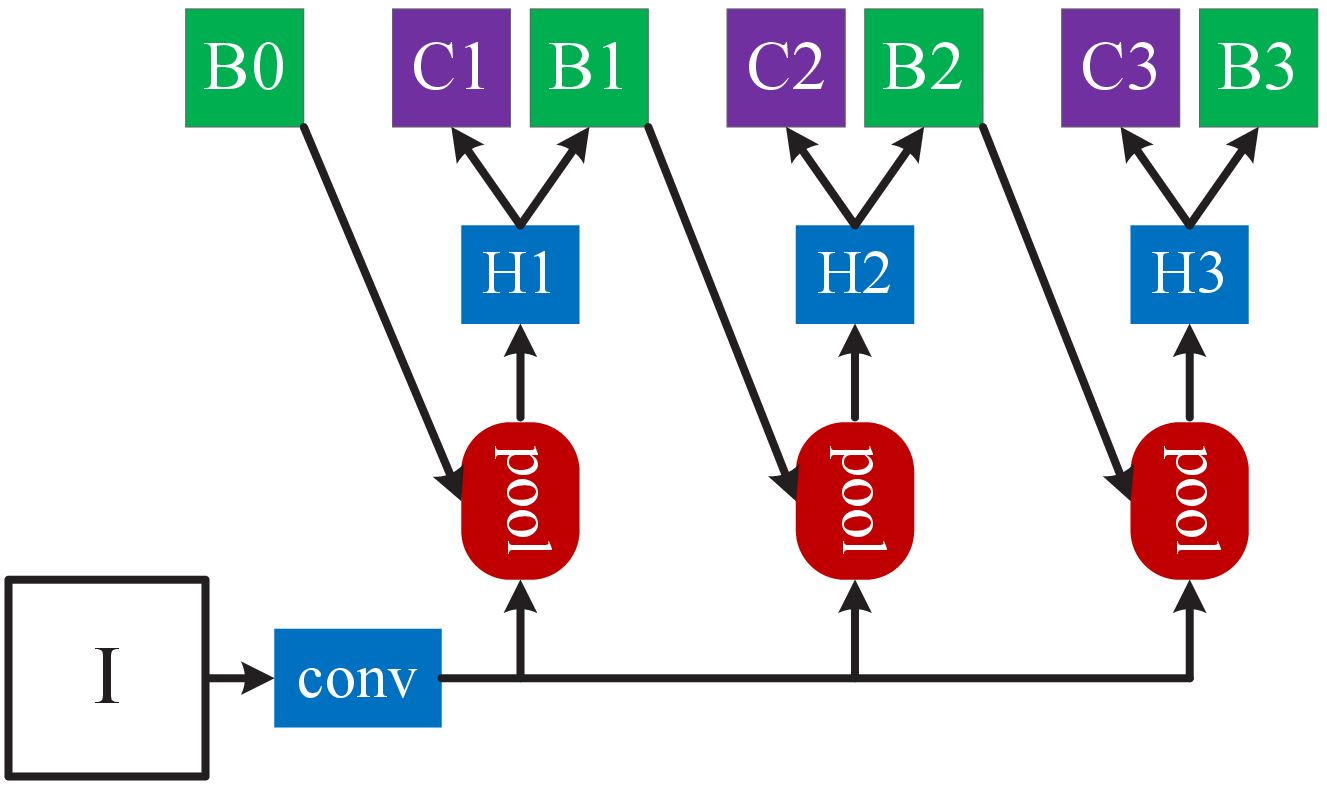,width=5.24cm,height=3.06cm}}{(b) Cascade R-CNN}
\end{minipage}
\hfill
\begin{minipage}[b]{.3\linewidth}
\centering
\centerline{\epsfig{figure=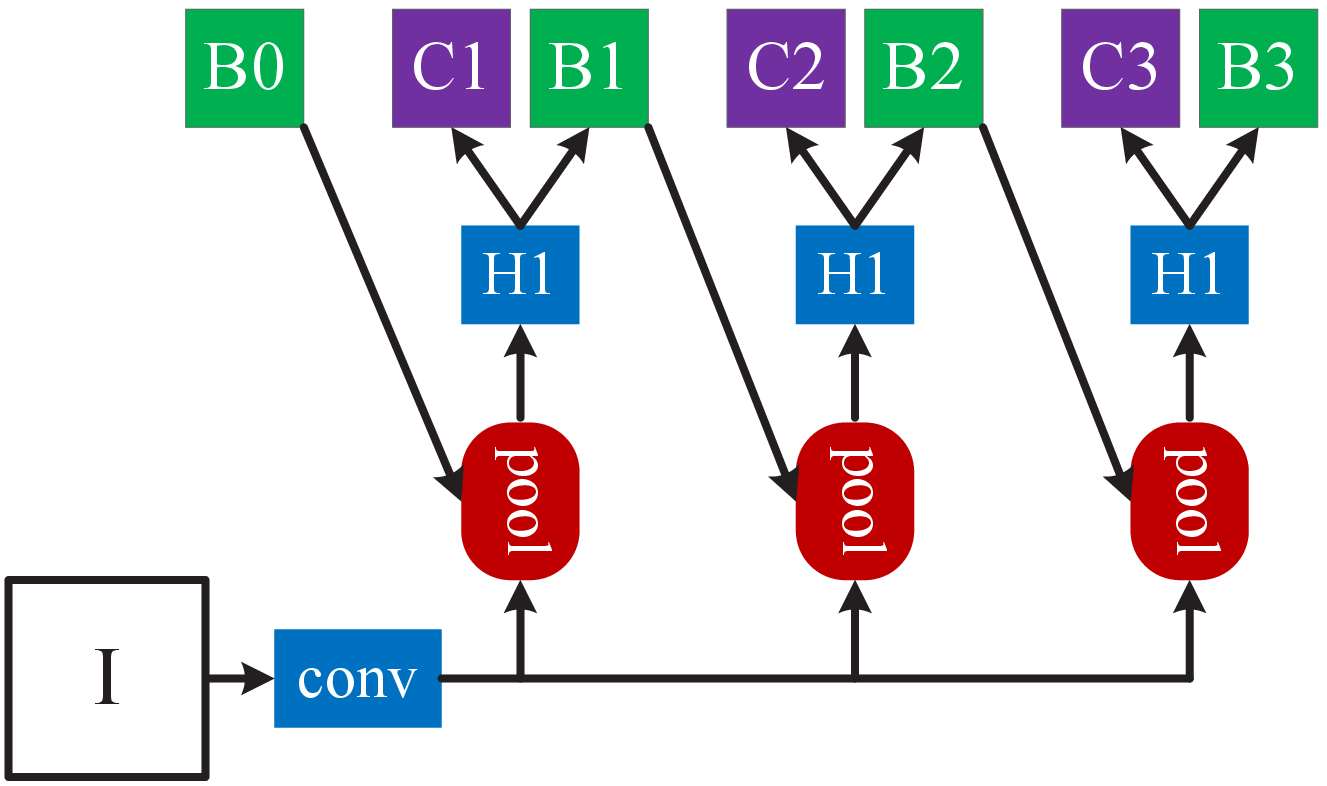,width=5.24cm,height=3.06cm}}{(c) Iterative BBox at inference}
\end{minipage}
\hfill
\begin{minipage}[b]{.21\linewidth}
\centering
\centerline{\epsfig{figure=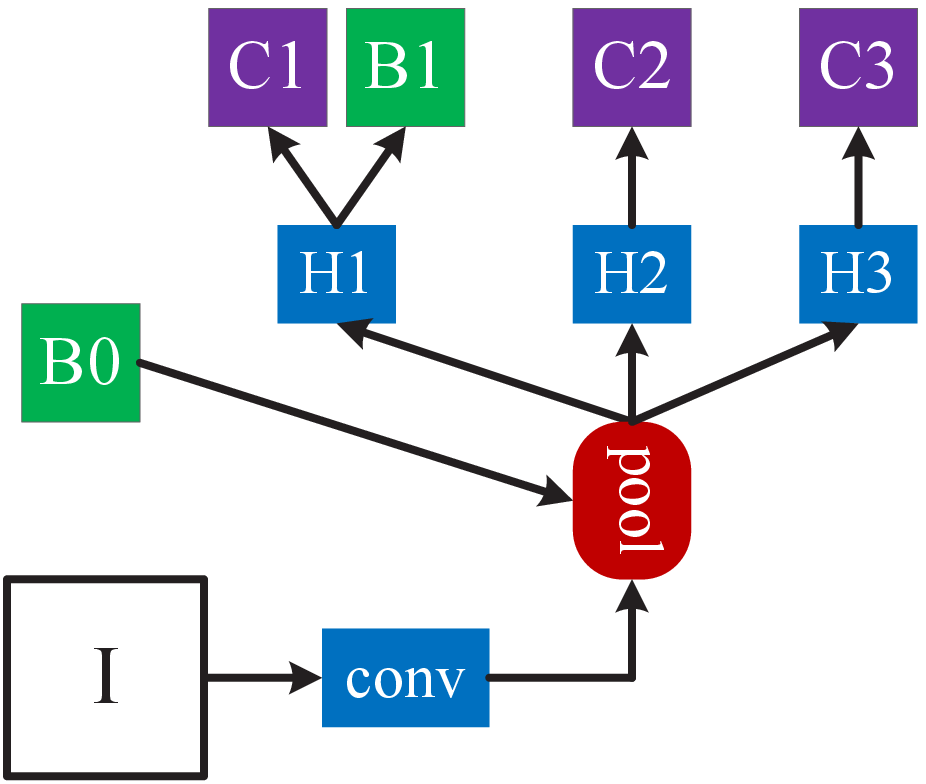,width=3.65cm,height=3.06cm}}{(d) Integral Loss}
\end{minipage}
\caption{The architectures of different frameworks. ``I'' is input image, ``conv'' backbone convolutions, ``pool'' region-wise feature extraction, ``H'' network head, ``B'' bounding box, and ``C'' classification. ``B0'' is proposals in all architectures.}
\label{fig:framework}
\end{figure*}

Some explorations in multi-stage object detection have also been proposed.
The multi-region detector of \cite{DBLP:conf/iccv/GidarisK15}
introduced \textit{iterative bounding box regression}, where a
R-CNN is applied several times to produce successively more accurate
bounding boxes. \cite{DBLP:conf/cvpr/YangYLL16,DBLP:conf/cvpr/GidarisK16,DBLP:conf/bmvc/GidarisK16} used a multi-stage procedure
to generate accurate proposals, which are forwarded to an accurate model
(e.g. Fast R-CNN). \cite{DBLP:conf/iccv/YooPLPK15,DBLP:conf/cvpr/NajibiRD16}
proposed an alternative procedure to localize objects sequentially. While
this is similar in spirit to the Cascade-RCNN, these
methods use the {\it same\/} regressor iteratively for accurate localization.
On the other hand,
\cite{DBLP:conf/cvpr/LiLSBH15,DBLP:journals/corr/OuyangWZW17}
embedded the classic cascade architecture of
\cite{DBLP:journals/ijcv/ViolaJ04} in an object detection network.
Finally, \cite{DBLP:conf/cvpr/DaiHS16} iterated between the
detection and segmentation tasks, to achieve improved instance
segmentation.

Upon publication of the conference version of this manuscript,
several works have pursued the idea behind Cascade R-CNN
\cite{DBLP:conf/eccv/JiangLMXJ18,DBLP:conf/eccv/LiuLHLC18,DBLP:journals/corr/abs-1803-08208,DBLP:journals/corr/abs-1901-07518}. \cite{DBLP:conf/eccv/LiuLHLC18,DBLP:journals/corr/abs-1803-08208} applied it to
single-shot object detectors, showing nontrivial improvements
for high quality single-shot detection, for general objects and
pedestrians, respectively. The IoU-Net \cite{DBLP:conf/eccv/JiangLMXJ18} explored
in greater detail high-quality localization,
achieving some gains over the Cascade R-CNN by cascading more bounding box
regression steps. \cite{he2018rethinking} showed it is possible to achieve state-of-the-art object detectors without ImageNet pretraining, with a help of the Cascade R-CNN. These works show that the Cascade R-CNN idea is robust
and applicable to various object detection architectures.
This suggests that it should continue to be useful despite future advances
in object detection.

\section{High Quality Object Detection}

In this section, we discuss the challenges of high quality
object detection.

\subsection{Object Detection}

While the ideas proposed in this work can be applied
to various detector architectures, we focus on the popular two-stage
architecture of the Faster R-CNN \cite{DBLP:conf/nips/RenHGS15}, shown in
Fig. \ref{fig:framework} (a). The first stage is a proposal sub-network,
in which the entire image is processed by a {\it backbone\/} network,
e.g. ResNet \cite{DBLP:conf/cvpr/HeZRS16}, and a proposal head (``H0'')
is applied to produce preliminary detection hypotheses, known as object
proposals. In the second stage, these
hypotheses are processed by a region-of-interest detection
sub-network (``H1''), denoted as a {\it detection head\/}. A final
classification score (``C'') and a bounding box (``B'') are assigned per
hypothesis. The entire detector is learned end-to-end, using a
multi-task loss with bounding box regression and classification
components.

\subsubsection{Bounding Box Regression}
\label{subsubsec:bbox}

A bounding box $\textbf{b}=(b_x,b_y,b_w,b_h)$ contains the four coordinates of
an image patch $\bf x$. Bounding box regression aims to regress a
candidate bounding box $\textbf{b}$ into a target
bounding box $\textbf{g}$, using a regressor $f({\bf x},\textbf{b})$. This is
learned from a training set
$(\textbf{g}_i,\textbf{b}_i)$, by minimizing the risk
\begin{equation}
  {\cal R}_{loc}[f] = \sum_{i}L_{loc}(f({\bf x}_i,\textbf{b}_i),\textbf{g}_i).
  \label{eq:bbrisk}
\end{equation}
As in Fast R-CNN \cite{DBLP:conf/iccv/Girshick15},
\begin{equation}
  L_{loc}({\bf a}, {\bf b}) = \sum_{i\in\{x,y,w,h\}} smooth_{L_1}(a_i-b_i)
\end{equation}
where
\begin{equation}
smooth_{L_1}(x)=\left\{
\begin{array}{cl}0.5x^2, &\quad\textrm{$|x|<1$}\\
|x|-0.5, &\quad\textrm{otherwise,} \end{array}\right.
\end{equation}
is the smooth $L_1$ loss function.
To encourage invariance to scale and location, $smooth_{L_1}$ operates on the
distance vector $\Delta=(\delta_x,\delta_y,\delta_w,\delta_h)$ defined by
\begin{equation}
\label{equ:delta}
\begin{array}{cl}\delta_x=(g_x-b_x)/b_w,\quad\delta_y=(g_y-b_y)/b_h\\
\delta_w=\log(g_w/b_w),\quad\delta_h=\log(g_h/b_h).\end{array}
\end{equation}
Since bounding box regression usually performs minor adjustments
on $\textbf{b}$, the numerical values of (\ref{equ:delta}) can be very
small. This usually makes the regression loss much smaller
than the classification loss. To improve the effectiveness of multi-task
learning, $\Delta$ is normalized by its mean and variance, e.g.
$\delta_x$ is replaced by
\begin{equation}
  \delta_x'= \frac{\delta_x-\mu_x}{\sigma_x}.
  \label{eq:norm}
\end{equation}
This is widely used in the literature \cite{DBLP:conf/nips/RenHGS15,DBLP:conf/eccv/CaiFFV16,DBLP:conf/nips/DaiLHS16,lin2017feature,he2017mask}.

\subsubsection{Classification}

The classifier is a function $h({\bf x})$ that assigns an image patch $\bf x$
to one of $M+1$ classes, where class $0$ contains background and the remaining
classes the objects to detect. $h({\bf x})$ is a $M+1$-dimensional estimate
of the posterior distribution over classes, i.e. $h_k({\bf x})=p(y=k|{\bf x})$,
where $y$ is the class label. Given a training set $({\bf x}_i, y_i)$, it is
learned by minimizing the classification risk
\begin{equation}
  {\cal R}_{cls}[h] = \sum_{i}L_{cls}(h({\bf x}_i),y_i),
  \label{eq:clsrisk}
\end{equation}
where
\begin{equation}
  L_{cls}(h({\bf x}),y) = -\log h_y({\bf x})
\end{equation}
is the cross-entropy loss.

\subsection{Detection Quality}

Consider a ground truth object of bounding
box ${\bf g}$ associated with class label $y$, and a detection hypothesis ${\bf x}$
of bounding box ${\bf b}$. Since a $\bf b$ usually includes an object and
some amount of background, it can be difficult to determine if a detection
is correct or not. This is usually addressed by the intersection over union
(IoU) metric
\begin{equation}
  IoU({\bf b}, {\bf g}) = \frac{{\bf b} \cap {\bf g}}{{\bf b} \cup
    {\bf g}}.
\end{equation}
If the IoU is above a threshold $u$, the patch is considered an example of
the class of the object of bounding box $\bf g$ and denoted ``positive''.
Thus, the class label of a hypothesis ${\bf x}$ is a function of $u$,
\begin{equation}
\label{equ:cls label}
y_u =\left\{
  \begin{array}{cl}
    y, &\quad\textrm{$IoU({\bf b},{\bf g})\geq{u}$}\\
    0, &\quad\textrm{otherwise.} \end{array}\right.
\end{equation}
If the IoU does not exceed the threshold for any object, $\bf x$ is
assigned to the background and denoted ``negative''.

Although there is no need to define positive/neagtive examples for the
bounding box regression task, an IoU threshold $u$ is also required to
select the set of samples
\begin{equation}
\mathcal{G}=\{(\textbf{g}_i,\textbf{b}_i) | IoU({\bf b}_i, {\bf g}_i)\geq{u}\}
\end{equation}
used to train the regressor. While the IoU thresholds used for the two tasks
do not have to be identical, this is usual in practice. Hence,
the IoU threshold $u$ defines the \textit{quality} of a detector.
Large thresholds encourage detected bounding boxes to be
tightly aligned with their ground truth counterparts.
Small thresholds reward detectors that produce loose bounding boxes,
of small overlap with the ground truth.

A main challenge of object detection is that, no matter the
choice of threshold, the detection setting is highly adversarial.
When $u$ is high, positives contain less background but it is difficult to
assemble large positive training sets. When $u$ is low, richer and
more diverse positive training sets are possible, but the trained
detector has little incentive to reject close false positives. In general,
it is very difficult to guarantee that a single
classifier performs uniformly well over all IoU levels. At inference,
since the majority of the hypotheses produced by a proposal detector,
e.g. RPN \cite{DBLP:conf/nips/RenHGS15} or selective
search \cite{DBLP:journals/ijcv/UijlingsSGS13}, have low quality, the detector
must be more discriminant for lower quality hypotheses. A standard
compromise between these conflicting requirements is to settle on $u=0.5$, which is used in almost {\it all} modern object detectors.
This, however, is a relatively low threshold, leading to low quality
detections that most humans consider close false positives, as shown in
Fig. \ref{fig:quality detection} (a).

\subsection{Challenges to High Quality Detection}
\label{subsec:high quality}

Despite the significant progress in object detection of the past few years,
few works attempted to address high quality detection. This is mainly due to
the following reasons.

First, evaluation metrics have historically placed greater emphasis on the
low quality detection regime. For performance evaluation, an IoU threshold
$u$ is used to determine whether a detection is a
success ($IoU({\bf b}, {\bf g})\geq{u}$) or failure
($IoU({\bf b}, {\bf g})<u$). Many object detection datasets, including
PASCAL VOC \cite{DBLP:journals/ijcv/EveringhamGWWZ10},
ImageNet \cite{DBLP:journals/ijcv/RussakovskyDSKS15},
Caltech Pedestrian\cite{DBLP:journals/pami/DollarWSP12}, etc., use $u=0.5$.
This is partly because these datasets were established a while ago, when
object detection performance was far from what it is today. However,
this loose evaluation standard is adopted even by relatively recent datasets,
such as WiderFace \cite{DBLP:conf/cvpr/YangLLT16}, or
CityPersons \cite{DBLP:conf/cvpr/ZhangBS17}. This is one of the main reasons
why performance has saturated for many of these datasets. Others,
such as COCO \cite{DBLP:conf/eccv/LinMBHPRDZ14} or
KITTI \cite{DBLP:conf/cvpr/GeigerLU12} use stricter evaluation metrics:
average precision at $u=0.7$ for car in KITTI, and mean average precision
across $u=[0.5:0.05:0.95]$ in COCO. While recent works have
focused on these less saturated datasets, most detectors are still designed
with the loose IoU threshold of $u=0.5$, associated with the low-quality
detection regime. In this work, we show that there is plenty of
room for improvement when a stricter evaluation metric, e.g. $u\geq{0.75}$, is
used and that it is possible to achieve significant improvements by
designing detectors specifically for the high quality regime.

\begin{figure}[!t]
\begin{minipage}[b]{.3\linewidth}
\centering
\centerline{\epsfig{figure=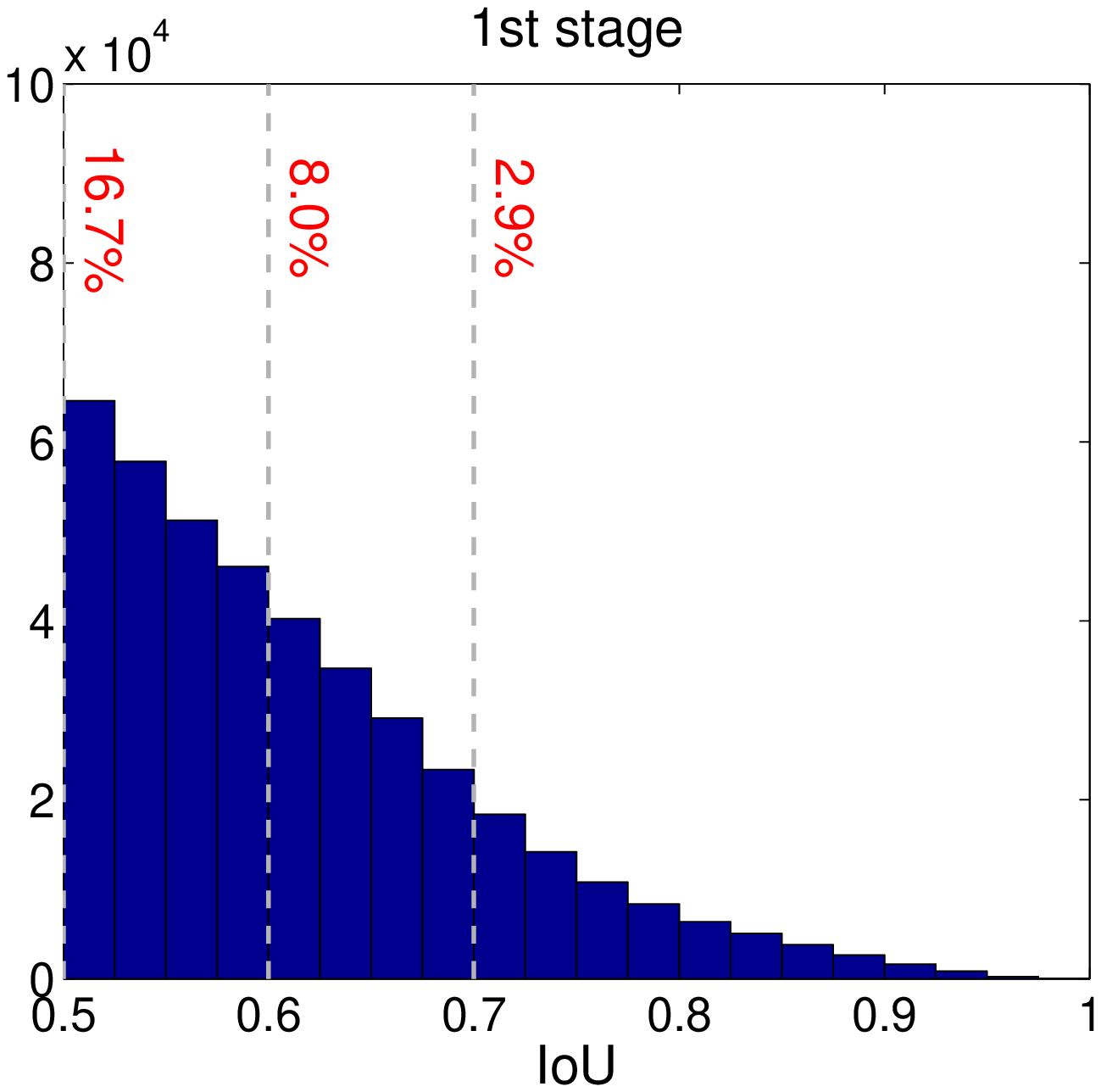,width=3.2cm,height=2.6cm}}
\end{minipage}
\hfill
\begin{minipage}[b]{.3\linewidth}
\centering
\centerline{\epsfig{figure=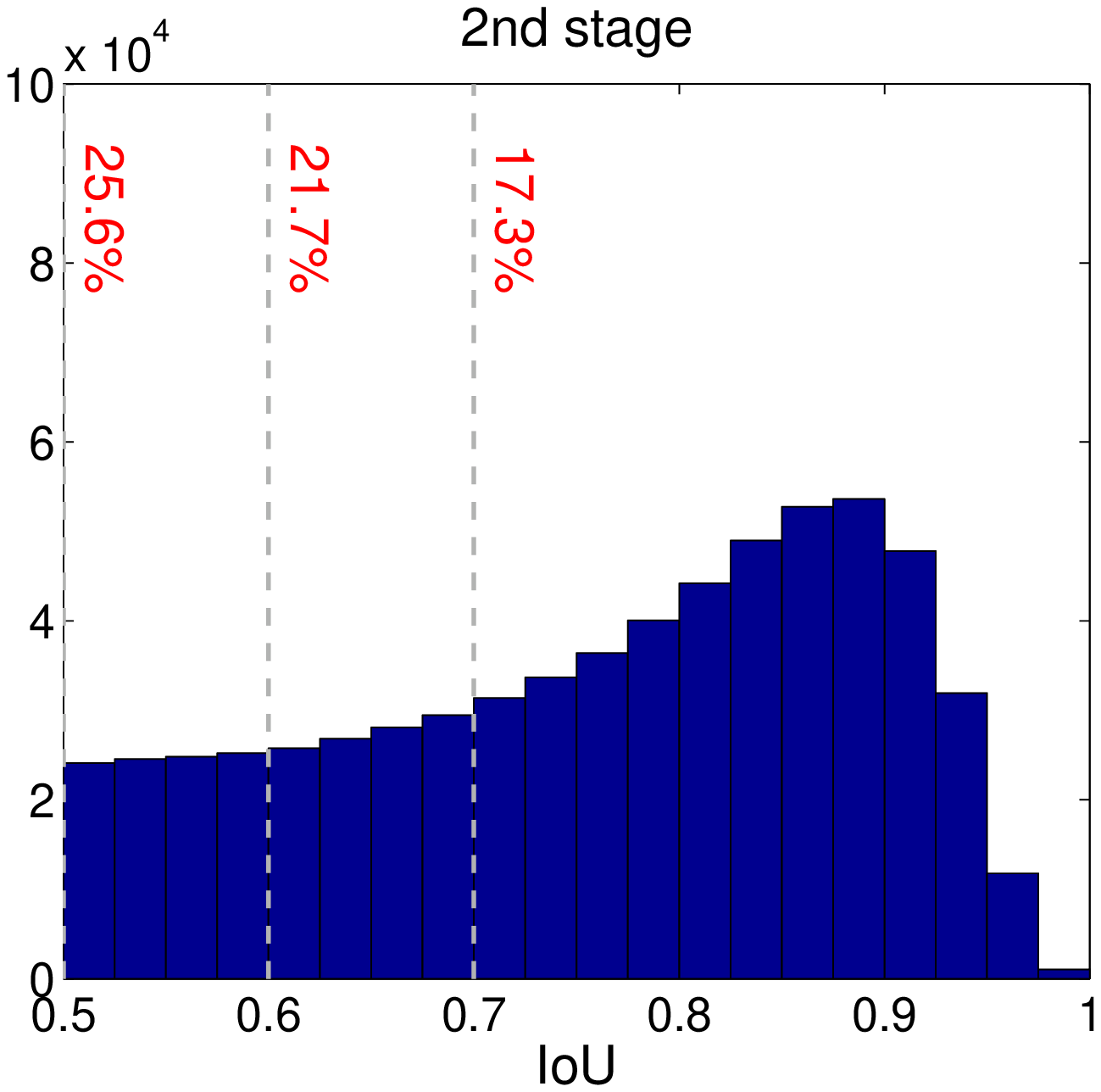,width=3.2cm,height=2.6cm}}
\end{minipage}
\hfill
\begin{minipage}[b]{.3\linewidth}
\centering
\centerline{\epsfig{figure=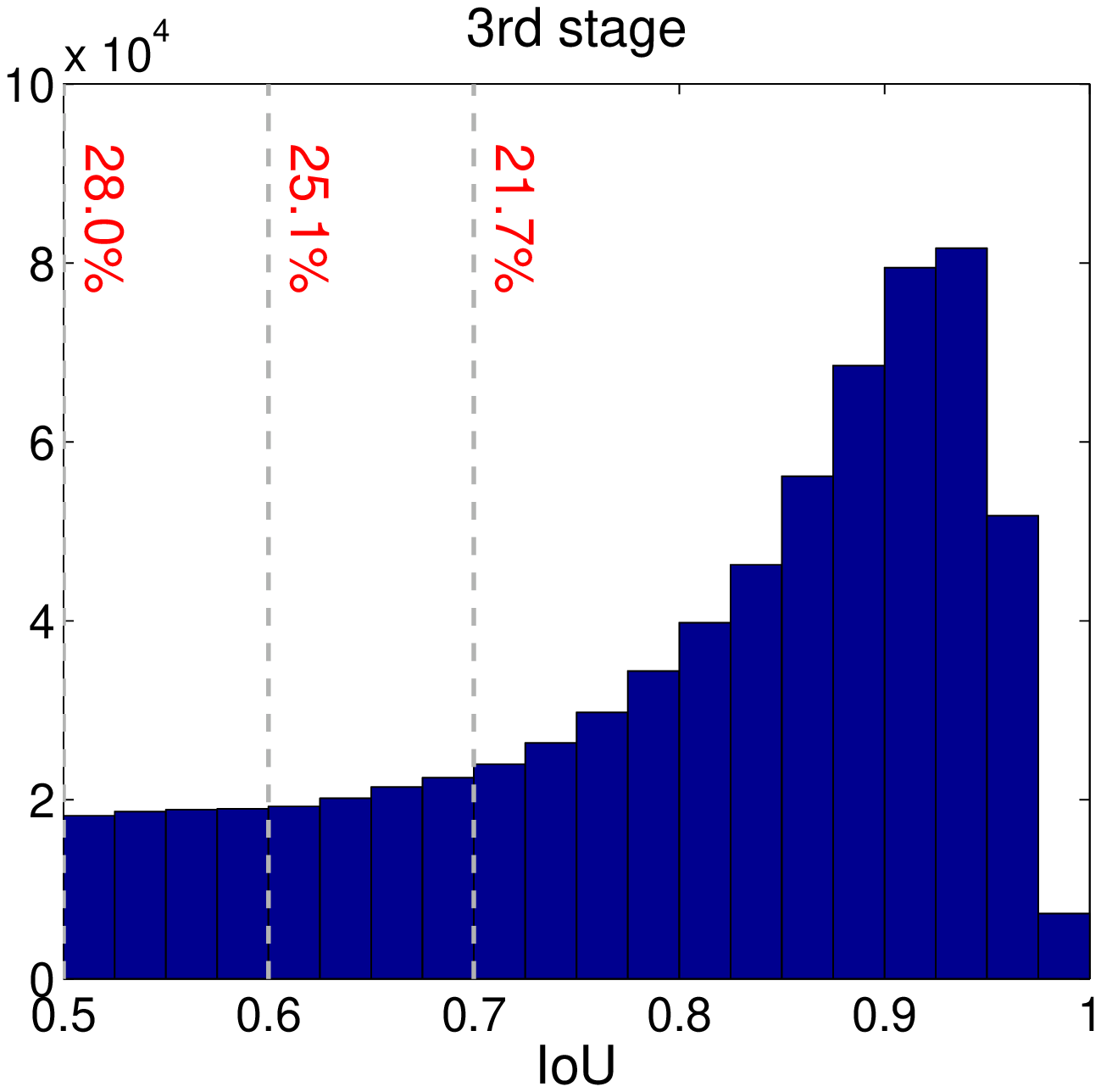,width=3.2cm,height=2.6cm}}
\end{minipage}
\caption{IoU histograms of training samples of each cascade stage. The
distribution of the 1st stage is the RPN output. Shown in red are the
percentage of positives for the corresponding IoU threshold.}
\label{fig:hist}
\end{figure}

Second, the design of high quality object detectors is
not a trivial generalization of existing approaches, due to the
paradox of high quality detection.
To beat the paradox, it is necessary to match the qualities of the
hypotheses generator and the object detector. In the literature,
there have been efforts to increase the quality of hypotheses, e.g. by
iterative bounding box regression \cite{DBLP:conf/cvpr/GidarisK16,DBLP:conf/bmvc/GidarisK16} or better RPN design
\cite{DBLP:conf/eccv/CaiFFV16,lin2017feature}, and some efforts to increase
the quality of the object detector, e.g. by using the integral loss on a set
of IoU thresholds \cite{DBLP:conf/bmvc/ZagoruykoLLPGCD16}. These attempts
fail to guarantee high quality detection because they consider
only one of the goals, missing the fact that the qualities of both
tasks need to be increased \textit{simultaneously}.
On one hand, raising the quality of the hypotheses has little benefit
if the detector remains of low quality, because the latter is not trained
to discriminate high quality from low quality  hypotheses.
On the other, if only the detector
quality is increased, there are too few high quality hypotheses for it
to classify, leading to no detection improvement. In fact,
because, as shown in Fig. \ref{fig:hist} (left),
the set of positive samples decreases quickly with $u$,
a high $u$ detector is prone to overfitting. Hence, a high $u$ detector can
easily overfit and perform worse than a low $u$
detector, as shown in Fig. \ref{fig:motivation} (c).

\section{Cascade R-CNN}

In this section we introduce the Cascade R-CNN detector.

\subsection{Architecture}

The architecture of the Cascade R-CNN is shown in Fig. \ref{fig:framework}
(b). It is a multi-stage extension of the Faster R-CNN architecture of
Fig. \ref{fig:framework} (a). In this work, we focus on the
the detection sub-network, simply adopting the
RPN \cite{DBLP:conf/nips/RenHGS15} of Fig. \ref{fig:framework} (a) for
proposal detection. However, the Cascade R-CNN is not limited to this
proposal mechanism, other choices should be possible. As discussed in the
section above, the goal is to increase the quality of hypotheses and
detector \textit{simultaneously}, to enable high quality object detection.
This is achieved with a combination of cascaded bounding box regression
and cascaded detection.

\subsection{Cascaded Bounding Box Regression}
\label{subsec:cascade bbox}

High quality hypotheses can be easily produced during training, where
ground truth bounding boxes are available, e.g. by sampling around the
ground truth. The difficulty is to produce high quality proposals at
inference, when ground truth is unavailable. This problem is addressed
with resort to cascaded bounding box regression.

As shown in Fig. \ref{fig:motivation} (a), a single regressor
cannot usually perform uniformly well over all quality levels.
However, as is commonly done for pose regression
\cite{DBLP:conf/cvpr/DollarWP10} or face
alignment \cite{DBLP:conf/cvpr/CaoWWS12,DBLP:conf/cvpr/XiongT13,yan2013learn},
the regression task can be decomposed into a sequence of simpler
steps. In the Cascade R-CNN detector, the idea is implemented
with a cascaded regressor with the architecture of
Fig. \ref{fig:framework} (b). This consists of a cascade
of {\it specialized\/} regressors
\begin{equation}
  f({\bf x},\textbf{b})=f_T\circ{f}_{T-1}\circ\cdots
  \circ{f}_1({\bf x},\textbf{b}),
\end{equation}
where $T$ is the total number of cascade stages. The key point
is that each regressor $f_t$ is optimized for the bounding box
distribution $\{\textbf{b}^t\}$ generated by the previous regressor, rather
than the initial distribution $\{\textbf{b}^1\}$. In this way, the hypotheses
are improved {\it progressively.\/}

\begin{figure}[!t]
\begin{minipage}[b]{.32\linewidth}
\centering
\centerline{\epsfig{figure=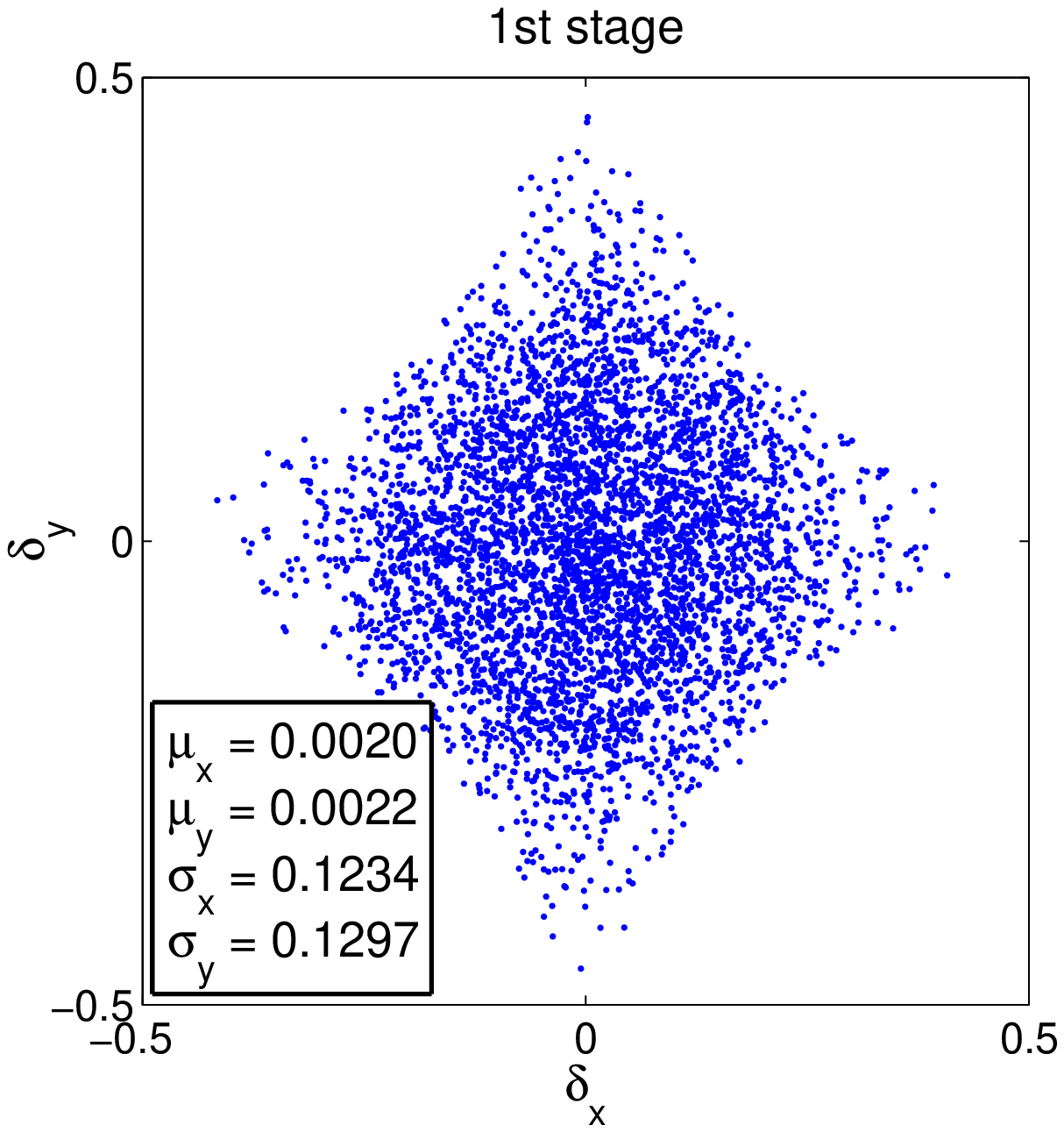,width=3cm,height=2.8cm}}
\end{minipage}
\hfill
\begin{minipage}[b]{.32\linewidth}
\centering
\centerline{\epsfig{figure=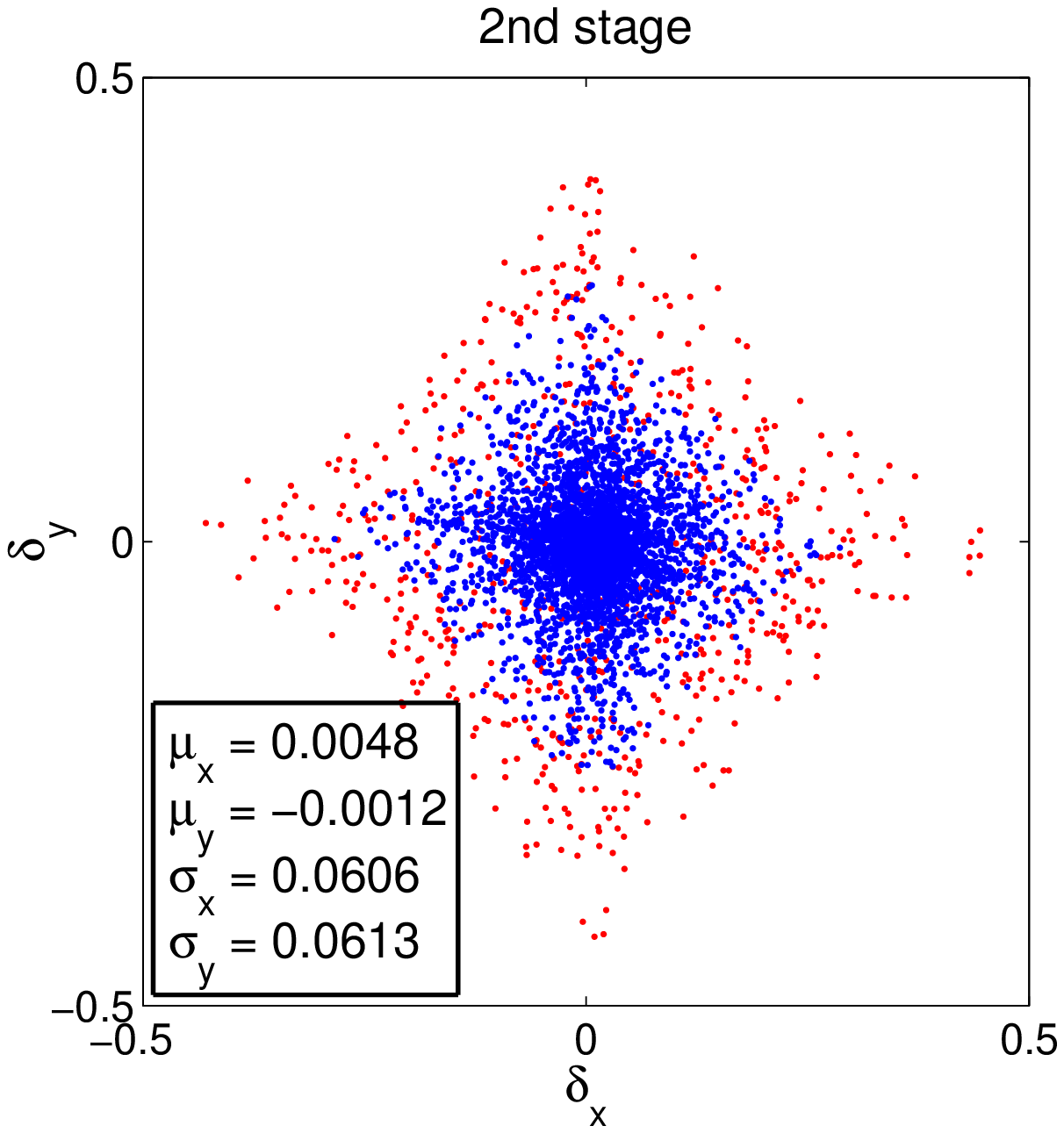,width=3cm,height=2.8cm}}
\end{minipage}
\hfill
\begin{minipage}[b]{.32\linewidth}
\centering
\centerline{\epsfig{figure=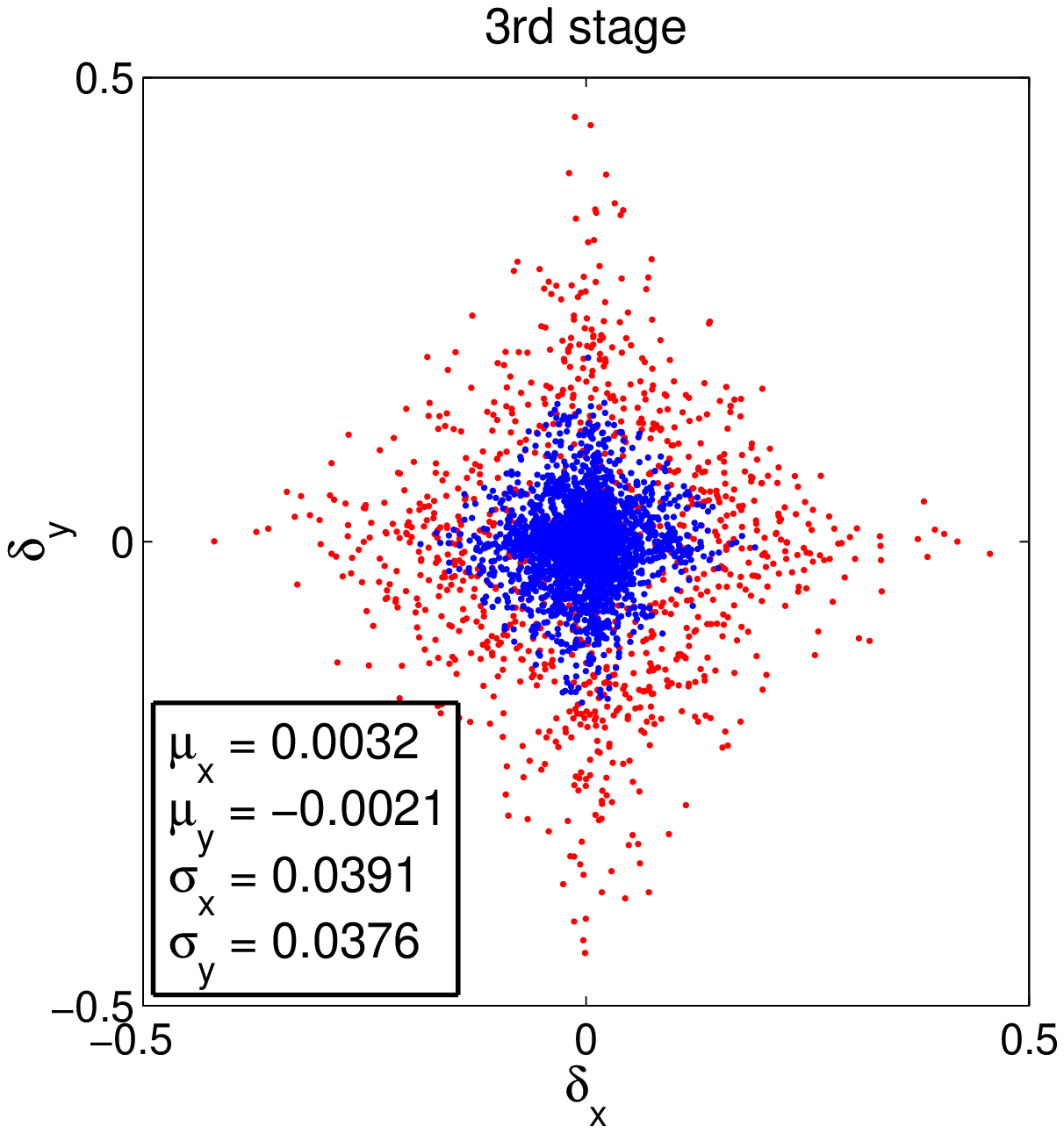,width=3cm,height=2.8cm}}
\end{minipage}\\
\hfill
\begin{minipage}[b]{.32\linewidth}
\centering
\centerline{\epsfig{figure=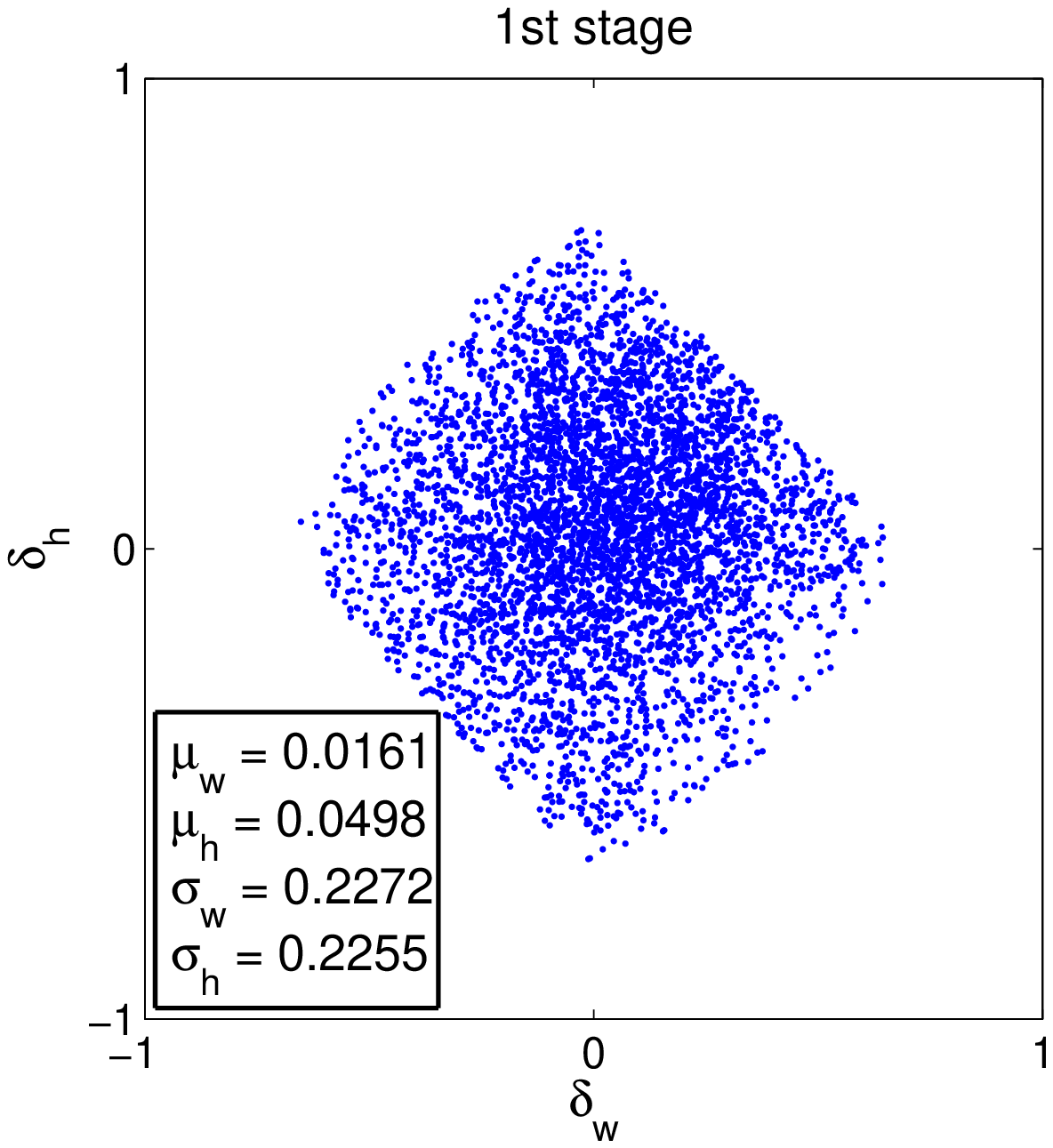,width=3cm,height=2.8cm}}
\end{minipage}
\hfill
\begin{minipage}[b]{.32\linewidth}
\centering
\centerline{\epsfig{figure=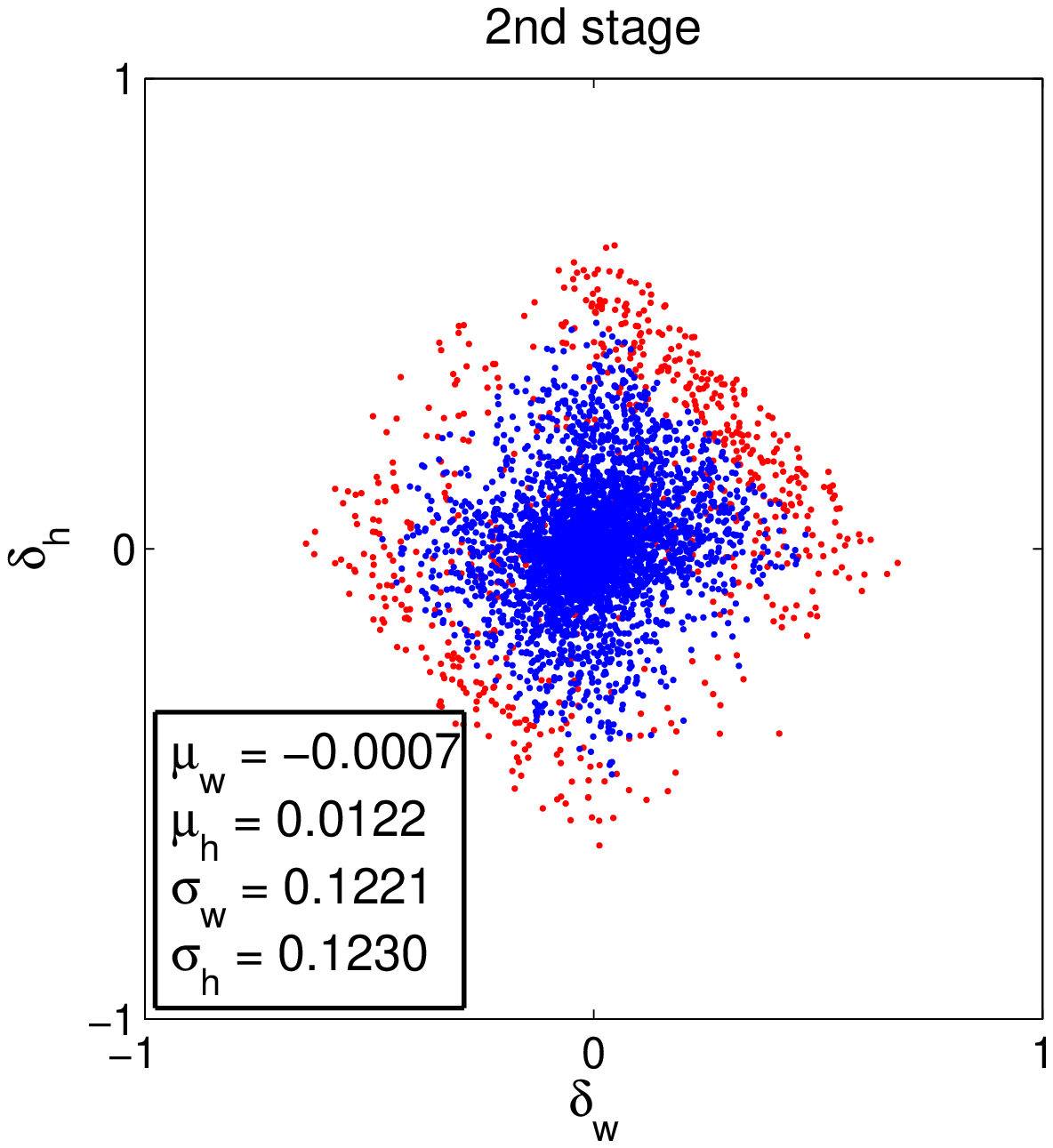,width=3cm,height=2.8cm}}
\end{minipage}
\hfill
\begin{minipage}[b]{.32\linewidth}
\centering
\centerline{\epsfig{figure=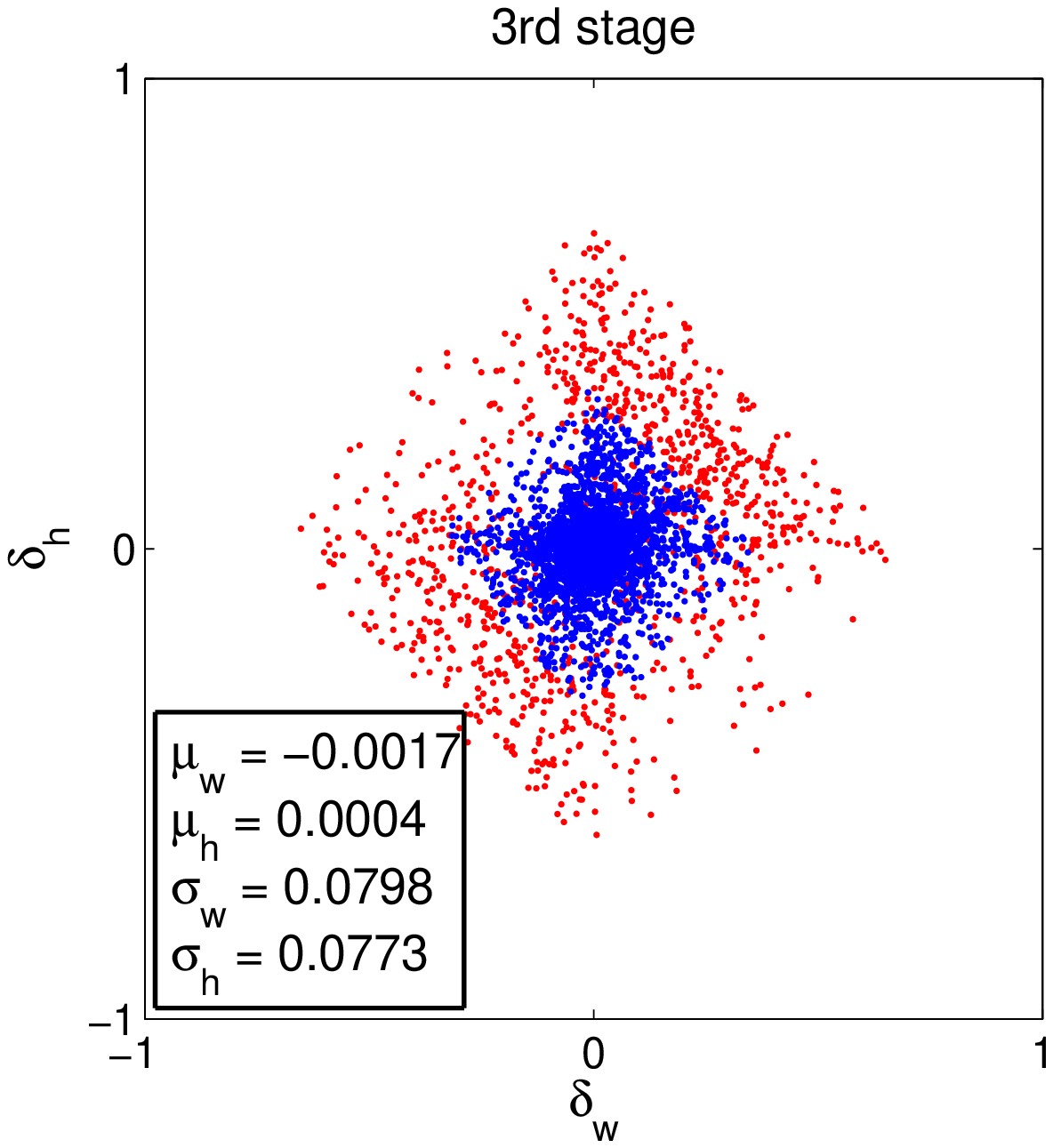,width=3cm,height=2.8cm}}
\end{minipage}
\caption{Distribution of the distance vector $\Delta$ of (\ref{equ:delta})
(without normalization) at different cascade stages. Top:
plot of $(\delta_x, \delta_y)$. Bottom: plot of $(\delta_w,\delta_h)$.
Red dots are outliers for the increasing IoU thresholds of
later stages, and the statistics shown are obtained after outlier removal.}
\label{fig:distribution}
\end{figure}

This is illustrated in
Fig. \ref{fig:distribution}, which presents the distribution of
the regression distance vector
$\Delta=(\delta_x,\delta_y,\delta_w,\delta_h)$ at different
cascade stages. Note that most hypotheses become closer to the ground truth
as they progress through the cascade. There are also some hypotheses
that fail to meet the stricter IoU criteria of the later
cascade stages. These are declared outliers and eliminated.
It should be noted that, as discussed in Section \ref{subsubsec:bbox},
$\Delta$ needs be mean/variance normalized, as in~(\ref{eq:norm}), for
effective multi-task learning. The mean and variance statistics
computed after this outlier removal step are used to
normalize $\Delta$ at each cascade stage. Our experiments show
that this implementation of cascaded bounding box regression generates
hypotheses of very high quality at both training and inference.

\subsection{Cascaded Detection}
\label{subsec:cascade}

As shown in the left of Fig. \ref{fig:hist}, the initial hypotheses
distribution produced by the RPN is heavily tilted towards low quality.
For example, only 2.9\% of examples are positive for an IoU threshold
$u=0.7$. This makes it difficult to train a high quality detector.
The Cascade R-CNN addresses the problem by using cascade regression as
a {\it resampling mechanism\/}. This is inspired by
Fig. \ref{fig:motivation} (a), where nearly all curves are
above the diagonal gray line, showing that a bounding box regressor trained
for a certain $u$ tends to produce bounding boxes of {\it higher\/} IoU.
Hence, starting from examples $\{({\bf x}_i,\textbf{b}_i)\}$, cascade
regression
successively resamples an example distribution $\{({\bf x}'_i,\textbf{b}'_i)\}$
of higher IoU. This enables the sets of positive
examples of the successive stages to keep a roughly {\it constant\/} size,
even when the detector quality $u$ is {\it increased\/}.
Figure \ref{fig:hist} illustrates this property, showing how the
example distribution tilts more heavily towards high quality examples after
each resampling step.

At each stage $t$, the R-CNN head includes a classifier $h_t$ and a
regressor $f_t$ optimized for the corresponding IoU threshold $u^t$,
where $u^t>u^{t-1}$. These are learned with loss
\begin{equation}
  L({\bf x}^t,g)=L_{cls}(h_t({\bf x}^t),y^t)+
  \lambda[y^t\geq{1}]L_{loc}(f_t({\bf x}^t,\textbf{b}^t),\textbf{g}),
\end{equation}
where $\textbf{b}^t=f_{t-1}({\bf x}^{t-1},\textbf{b}^{t-1})$, $\bf g$ is the
ground truth object for ${\bf x}^t$, $\lambda=1$ the trade-off coefficient,
$y^t$ is the label of ${\bf x}^t$ under the $u^t$ criterion, according to
(\ref{equ:cls label}), $[\cdot]$ is the indicator function. Note that
the use of $[\cdot]$ implies that the IoU threshold $u$ of bounding box
regression is identical to that used for classification. This cascade learning has
three important consequences for detector training. First, the
potential for overfitting at large IoU thresholds $u$ is reduced, since
positive examples become plentiful at all stages (see Fig.~\ref{fig:hist}).
Second, detectors of deeper stages are optimal for higher IoU
thresholds. Third, because some outliers are removed as the IoU threshold
increases (see Fig. \ref{fig:distribution}), the learning effectiveness of bounding
box regression increases in the later stages. This simultaneous improvement
of hypotheses and detector quality enables the Cascade R-CNN to beat
the paradox of high quality detection. At inference, the same cascade is applied.
The quality of the hypotheses is improved sequentially, and higher quality
detectors are only required to operate on higher quality hypotheses,
for which they are optimal. This enables the high quality object detection
results of Fig. \ref{fig:quality detection} (b), as suggested by Fig. \ref{fig:motivation}.

\begin{figure*}[!t]
\begin{minipage}[b]{.15\linewidth}
\centering
\centerline{\epsfig{figure=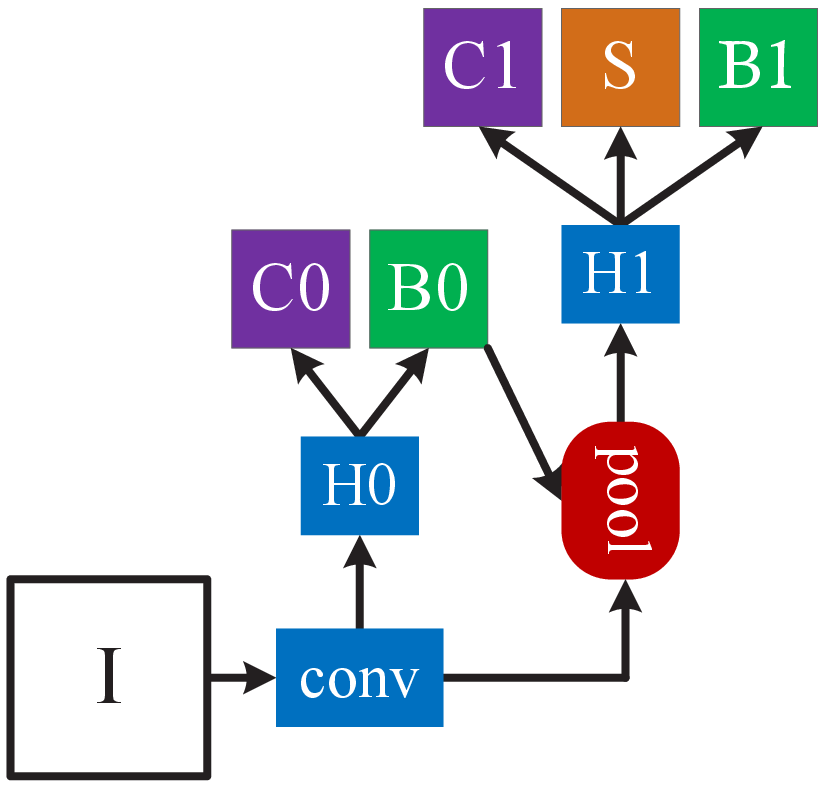,width=2.93cm,height=2.78cm}}{(a)}
\end{minipage}
\hfill
\begin{minipage}[b]{.25\linewidth}
\centering
\centerline{\epsfig{figure=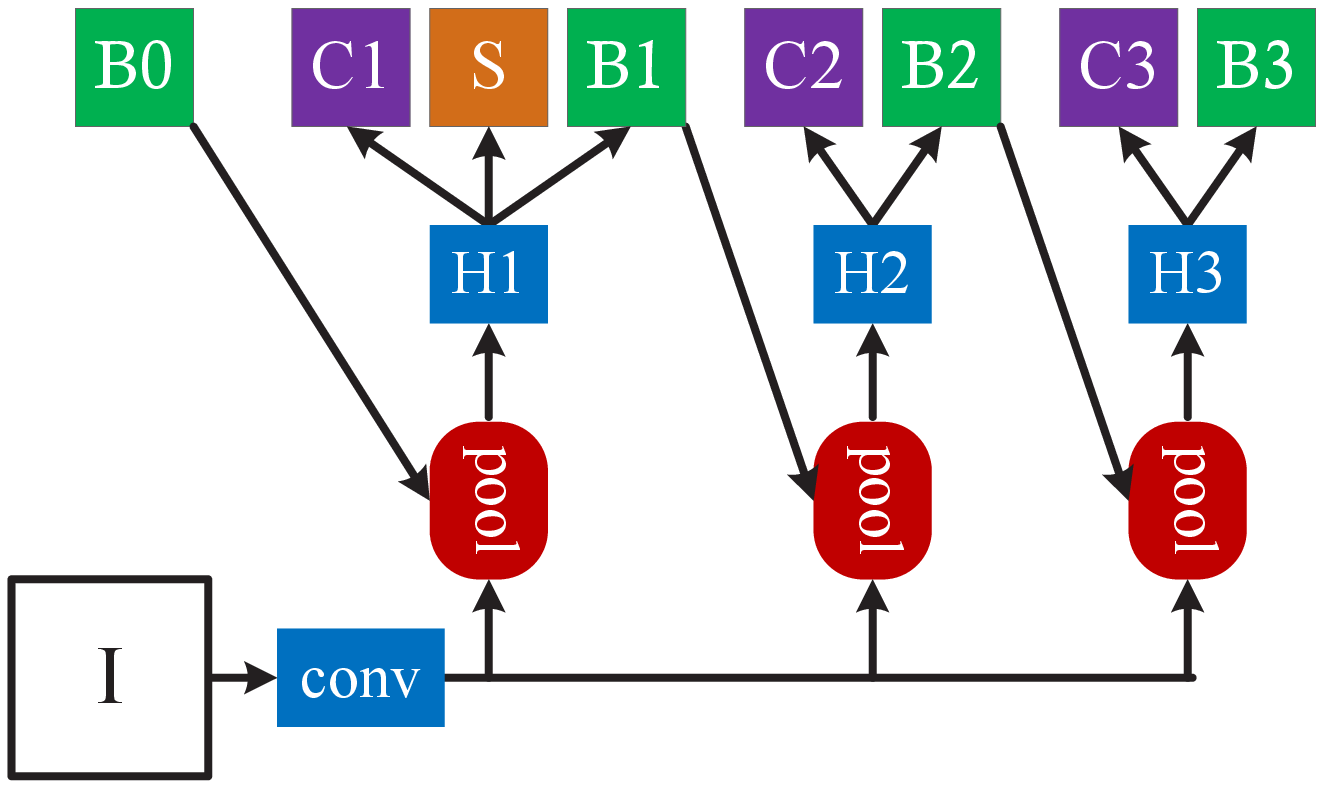,width=4.73cm,height=2.78cm}}{(b)}
\end{minipage}
\hfill
\begin{minipage}[b]{.26\linewidth}
\centering
\centerline{\epsfig{figure=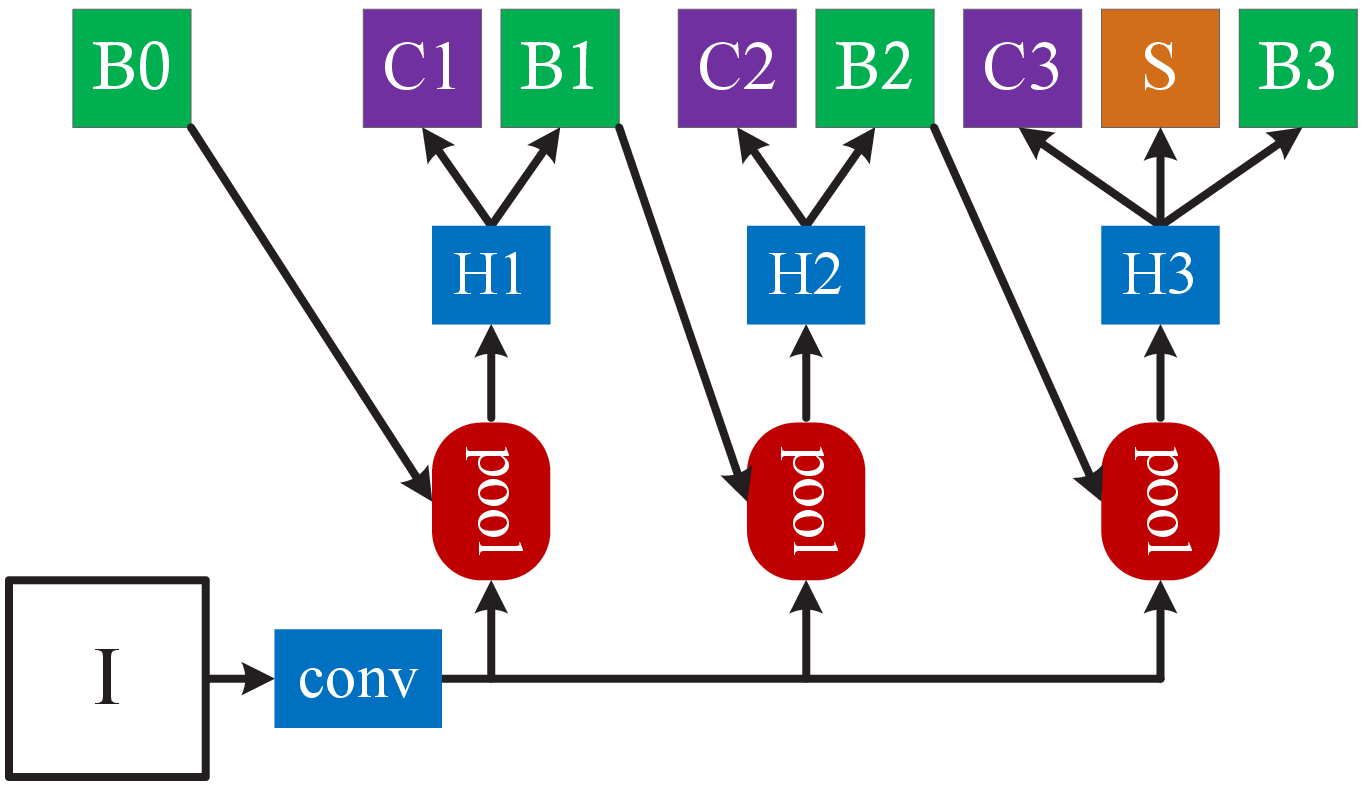,width=4.89cm,height=2.78cm}}{(c)}
\end{minipage}
\hfill
\begin{minipage}[b]{.29\linewidth}
\centering
\centerline{\epsfig{figure=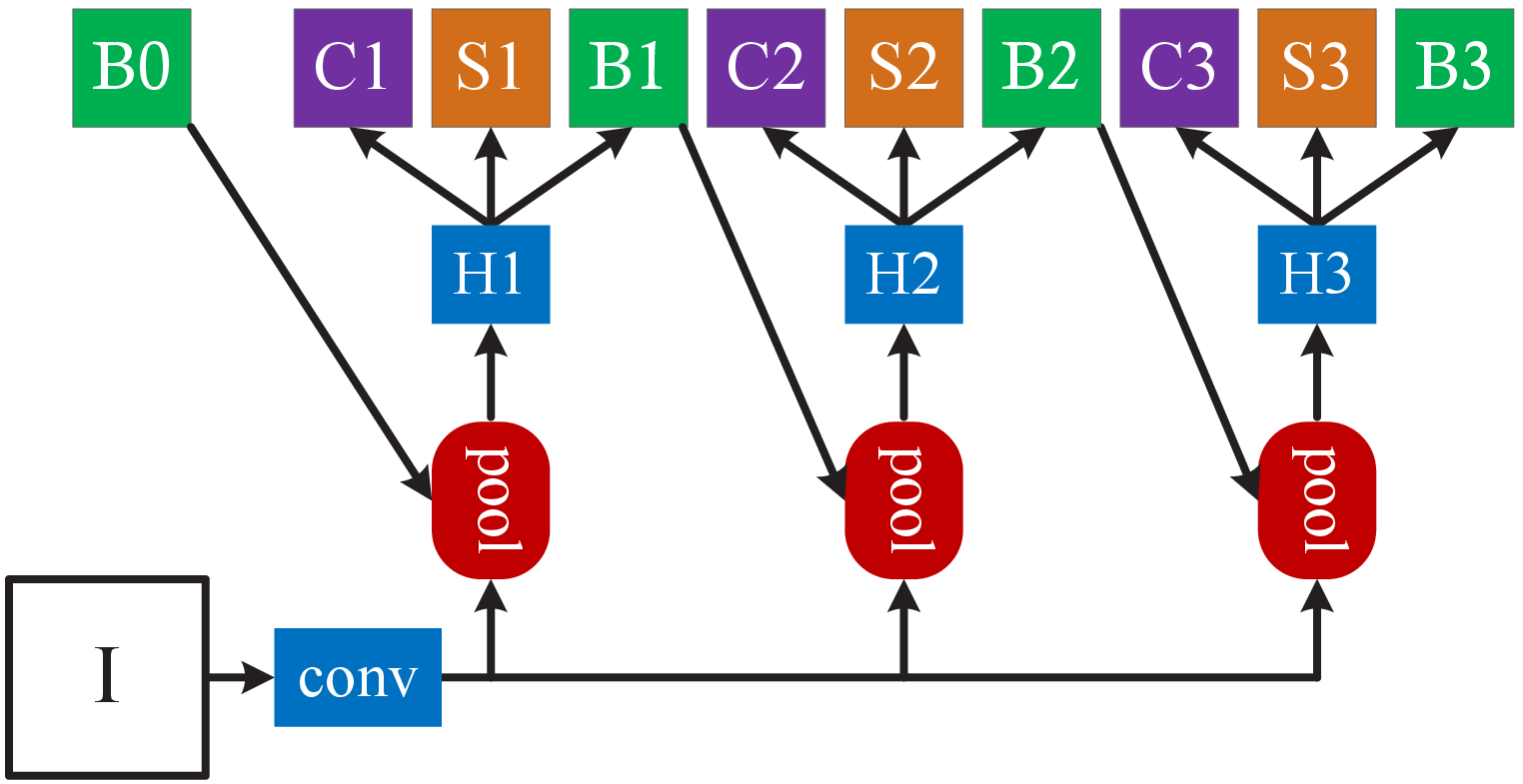,width=5.46cm,height=2.78cm}}{(d)}
\end{minipage}
\caption{Architectures of the Mask R-CNN (a) and three Cascade Mask R-CNN strategies
for instance segmentation (b)-(d). Beyond the definitions of
Fig. \ref{fig:framework}, ``S'' denotes a segmentation branch.
Note that segmentations branches do not necessarily share heads with the
detection branch.}
\label{fig:maskframework}
\end{figure*}

\subsection{Differences from Previous Works}

The Cascade R-CNN has similarities to previous works using \textit{iterative bounding box regression} and \textit{integral loss} for detection. There are, however, important differences.

\vspace{0.2cm}
\noindent{\bf Iterative Bounding Box Regression:}
Some works \cite{DBLP:conf/iccv/GidarisK15,DBLP:conf/bmvc/GidarisK16,DBLP:conf/cvpr/HeZRS16} have previously argued that the use of a single
bounding box regressor $f$ is insufficient for accurate localization.
These methods apply $f$ iteratively, as a post-processing step
\begin{equation}
\label{equ:iterative bbox}
f'({\bf x},\textbf{b})=f\circ{f}\circ\cdots\circ{f}({\bf x},\textbf{b}),
\end{equation}
that refines a bounding box $\textbf{b}$. This is called
\textit{iterative bounding box regression} and denoted as
\textit{iterative BBox}. It can be implemented with the inference architecture
of Fig. \ref{fig:framework} (c) where all heads are {\it identical\/}.
Note that this is only for inference, as training is identical to that of a
two-stage object detector, e.g. the Faster R-CNN of Fig.
\ref{fig:framework} (a) with $u=0.5$.
This approaches ignores two problems. First, as shown in Fig.
\ref{fig:motivation}, a regressor $f$ trained at $u=0.5$ is
suboptimal for hypotheses of higher IoUs. It actually {\it degrades\/}
bounding box accuracy for IoUs larger than $0.85$. Second, as shown in
Fig. \ref{fig:distribution}, the distribution of bounding boxes changes
significantly after each iteration. While the regressor is optimal for the
initial distribution it can be quite suboptimal after that. Due to these
problems, \textit{iterative BBox} requires a fair amount of human
engineering, in the form of proposal accumulation, box voting,
etc, and has somewhat unreliable gains \cite{DBLP:conf/iccv/GidarisK15,DBLP:conf/bmvc/GidarisK16,DBLP:conf/cvpr/HeZRS16}. Usually, there is no benefit beyond applying $f$ twice.

The Cascade R-CNN differs from \textit{iterative BBox} in several ways. First,
while \textit{iterative BBox} is a post-processing procedure used to
improve bounding boxes, the Cascade R-CNN uses cascade regression as
a {\it resampling\/} mechanism that changes the distribution of hypotheses
processed by the different stages. Second, because cascade regression
is used at both training and inference, there is no discrepancy between
training and inference distributions. Third,
the multiple specialized regressors $\{f_T,f_{T-1},\cdots,f_1\}$ are
optimal for the {\it resampled distributions\/} of the different stages.
This is unlike the single $f$ of (\ref{equ:iterative bbox}),
which is only optimal for the initial distribution. Our experiments
show that the Cascade R-CNN enables more precise localization than
that possible with \textit{iterative BBox}, and requires no human
engineering.

\vspace{0.2cm}
\noindent{\bf Integral Loss:}
\cite{DBLP:conf/bmvc/ZagoruykoLLPGCD16} proposed an ensemble of classifiers
with the architecture of Fig. \ref{fig:framework} (d) and trained
with the integral loss. This is a loss
\begin{equation}
  \label{equ:intloss}
  L_{cls}(h({\bf x}),y)=\sum_{u\in{U}}L_{cls}(h_u({\bf x}),y_u)
\end{equation}
that targets various quality levels, defined by a set of IoU thresholds
$U=\{0.5,0.55,\cdots,0.75\}$, chosen to fit the
evaluation metric of the COCO challenge.

The Cascade R-CNN differs from this detector in several ways. First,
\eqref{equ:intloss} fails to address the problem that
the various loss terms operate on different numbers of
positives. As shown on Fig. \ref{fig:hist} (left), the set of
positive samples decreases quickly with $u$. This is particularly
problematic because it makes the high quality classifiers very prone to
overfitting. On the other hand, as shown in Fig. \ref{fig:hist}, the
resampling of the Cascade R-CNN produces a nearly constant number of positive
examples as the IoU threshold $u$ increases.
Second, at inference, the high quality classifiers are required to process
proposals of overwhelming low quality, for which they are not
optimal. This is unlike the higher quality detectors of the Cascade R-CNN,
which are only required to operate on higher quality hypotheses.
Third, the integral loss is designed to fit the COCO metrics and,
by definition, the classifiers are ensembled at inference. The Cascade R-CNN
aims to achieve high quality detection, and the high quality detector itself
in the last stage can obtain the state-of-the-art detection
performance. Due to all this,
the \textit{integral loss} detector of Fig. \ref{fig:framework} (d) usually fails to
outperform the vanilla detector of
Fig. \ref{fig:framework} (a), for most quality levels. This is unlike the
Cascade R-CNN, which can have significant improvements.

\section{Instance Segmentation}
\label{sec:mask}

Instance segmentation has become popular in the recent
past \cite{DBLP:conf/cvpr/DaiHS16,he2017mask,liu2018path}. It aims to
predict pixel-level segmentation for each instance, in addition to determining
its object class. This is more difficult than object detection, which only
predicts a bounding box (plus class) per instance. In general, instance
segmentation is implemented in addition to object detection, and a stronger
object detector usually leads to improved instance segmentation.
The most popular instance segmentation method is arguably the
Mask R-CNN \cite{he2017mask}. Like the Cascade R-CNN, it is a variant
on the two-stage detector. In this section, we extend
the Cascade R-CNN architecture to the instance segmentation task, by
adding a segmentation branch similar to that of the Mask R-CNN.

\subsection{Mask R-CNN}

The Mask R-CNN \cite{he2017mask} extends the Faster R-CNN by adding a
segmentation branch in parallel to the existing detection branch during training. It has the architecture of
Fig. \ref{fig:maskframework} (a). The training instances are the positive
examples also used to train the detection task. At inference, object detections
are complemented with segmentation masks, for all detected
objects.

\subsection{Cascade Mask R-CNN}

In the Mask R-CNN, the segmentation branch is inserted in parallel to the
detection branch. However, the Cascade R-CNN has multiple
detection branches. This raises the questions of 1) where to add the
segmentation branch and 2) how many segmentation branches to add.
We consider three strategies for mask prediction in the Cascade R-CNN.
The first two strategies address the first question, adding a single mask prediction head at either the first or last stage of the Cascade R-CNN, as shown in
Fig. \ref{fig:maskframework} (b) and (c), respectively.
Since the instances used to train the segmentation branch are
the positives of the detection branch, their number varies in these two
strategies. As shown in Fig. \ref{fig:hist}, placing the
segmentation head later on the cascade leads to more examples.
However, because segmentation is a pixel-wise operation, a large number
of highly overlapping instances is not necessarily as helpful as for object
detection, which is a patch-based operation. The third strategy
addresses the second question, adding a segmentation branch to each cascade
stage, as shown in Fig. \ref{fig:maskframework} (d).
This maximizes the diversity of samples used to learn the mask prediction
task.

At inference time, all three strategies predict the segmentation masks on the
patches produced by the final object detection stage, irrespective
of the cascade stage on which the segmentation mask is implemented and
how many segmentation branches there are. The final mask prediction is
obtained from the single segmentation branch for the architectures of
Fig. \ref{fig:maskframework} (b) and (c), and from the ensemble of three
segmentation branches for the architecture of Fig. \ref{fig:maskframework} (d).
Our experiments show that these
architectures of the Cascade Mask R-CNN outperform the Mask R-CNN.

\section{Experimental Results}

In this section, we present an extensive evaluation of the Cascade R-CNN
detector.

\subsection{Experimental Set-up}

Experiments were performed over multiple datasets and baseline network
architectures.

\subsubsection{Datasets}

The bulk of the experiments was performed on MS-COCO
2017 \cite{DBLP:conf/eccv/LinMBHPRDZ14}, which contains $\sim$118k images
for training, 5k for validation (\texttt{val}) and $\sim$20k for testing
without provided annotations (\texttt{test-dev}). The COCO average
precision (AP) measure averages AP across IoU thresholds from 0.5 to 0.95,
with an interval of 0.05. It measures detection performance
at various qualities, encouraging high quality detection results, as
discussed in Section \ref{subsec:high quality}. All models were trained on
the COCO training set and evaluated on the \texttt{val} set.
Final results are also reported on the \texttt{test-dev} set for fair
comparison with the state-of-the-art. To assess the robustness
and generalization ability of the Cascade R-CNN, experiments were also
performed on Pascal VOC \cite{DBLP:journals/ijcv/EveringhamGWWZ10},
KITTI \cite{DBLP:conf/cvpr/GeigerLU12},
CityPersons \cite{DBLP:conf/cvpr/ZhangBS17} and
WiderFace \cite{DBLP:conf/cvpr/YangLLT16}. Instance
segmentation was also evaluated on COCO, using the same evaluation metrics
as object detection. The only difference is that the IoU is computed with
respect to the mask rather than a bounding box.

\subsubsection{Implementation Details}

All regressors are class agnostic for simplicity. All Cascade R-CNN
detection stages have the same architecture, which is the detection head of
the baseline detector. Unless otherwise noted, the Cascade R-CNN is
implemented with four stages: one RPN and three detection heads with
thresholds $U=\{0.5,0.6,0.7\}$. The sampling of the first detection stage
follows \cite{DBLP:conf/iccv/Girshick15,DBLP:conf/nips/RenHGS15}. In
subsequent stages, resampling is implemented by using \textit{all}
the regressed outputs from the previous stage, as discussed in
Section \ref{subsec:cascade}. No data augmentation was used except standard horizontal image flipping. Inference was performed at a single image
scale, with no further bells and whistles. All baseline detectors were
reimplemented with Caffe \cite{DBLP:conf/mm/JiaSDKLGGD14}, using
the same codebase, for fair comparison. Some experiments with
the FPN and Mask R-CNN baselines were implemented on the Detectron platform.

\subsubsection{Baseline Networks}
\label{subsubsec:baseline}

To test the versatility of the Cascade R-CNN, experiments were performed with
multiple popular baselines: Faster R-CNN and
MS-CNN \cite{DBLP:conf/eccv/CaiFFV16} with VGG-Net \cite{DBLP:journals/corr/SimonyanZ14a} backbone,
R-FCN \cite{DBLP:conf/nips/DaiLHS16} and FPN \cite{lin2017feature} with
ResNet backbones \cite{DBLP:conf/cvpr/HeZRS16}, for the task of object detection, and Mask R-CNN \cite{he2017mask} with ResNet backbones for instance segmentation.
These baselines have a wide range of performances. Unless noted, their
default settings were used. End-to-end training was used instead of
multi-step training.

\vspace{0.2cm}
\noindent{\bf Faster R-CNN:}
the network head has two fully connected layers. To reduce parameters,
\cite{DBLP:conf/nips/HanPTD15} was used to prune less important connections.
2048 units were retained per fully connected layer and dropout layers were
removed. These changes have negligible effect on detection performance.
Training started with a learning rate of 0.002, which was reduced by a factor
of 10 at 60k and 90k iterations, and stopped at 100k iterations, on 2
synchronized GPUs, each holding 4 images per iteration. 128 RoIs were used
per image.

\vspace{0.2cm}
\noindent{\bf R-FCN:}
the R-FCN adds a convolutional, a bounding box regression, and a
classification layer to the ResNet. For this baseline, all
Cascade R-CNN heads have this structure. Online hard negative
mining \cite{DBLP:conf/cvpr/ShrivastavaGG16}
was not used. Training started with a learning rate of 0.003, which was
decreased by a factor of 10 at 160k and 240k iterations, and stopped at 280k
iterations, on 4 synchronized GPUs, each holding one image per iteration.
256 RoIs were used per image.

\vspace{0.2cm}
\noindent{\bf FPN:}
since official source code was not publicly available for the FPN when we
performed our original experiments \cite{cai18cascadercnn},
the implementation details were somewhat different from those later made
available in the Detectron implementation.
RoIAlign \cite{he2017mask} was used for a stronger baseline. This is denoted
as FPN+ and was used in all ablation studies, with the ResNet-50 as a
backbone as usual. Training used a learning rate of 0.005 for 120k iterations and
0.0005 for the next 60k iterations, on 8 synchronized GPUs,
each holding one image per iteration. 256 RoIs were used per image.
We have also reimplemented the Cascade R-CNN of FPN on Detectron platform, when it is
publicly available.

\vspace{0.2cm}
\noindent{\bf MS-CNN:}
the MS-CNN \cite{DBLP:conf/eccv/CaiFFV16} is a popular multi-scale object
detector for specific object categories, e.g. vehicle, pedestrian, face, etc.
It was used as baseline detector for experiments on KITTI, CityPersons and WiderFace.
For this baseline, the Cascade R-CNN adopted the same two-step training
strategy of the MS-CNN: proposal sub-network trained
first and then joint end-to-end training. All detection heads were only added at
the second step, where the learning rate was initially 0.0005,
decreased by a factor of 10 at 10k and 20k iterations and stopped at
25k iterations, on one GPU of batch size 4 images.

\vspace{0.2cm}
\noindent{\bf Mask R-CNN:}
the Mask R-CNN was used as baseline for instance segmentation. The default
Detectron implementation was adopted, using the \texttt{1x} learning schedule.
Training started with a learning rate of 0.02, which was reduced by a factor
of 10 at 60k and 80k iterations, and stopped at 90k iterations, on 8
synchronized GPUs, each holding 2 images per iteration. 512 RoIs were
used per image.

\subsection{Quality Mismatch}

\begin{figure}[!t]
\begin{minipage}[b]{.48\linewidth}
\centering
\centerline{\epsfig{figure=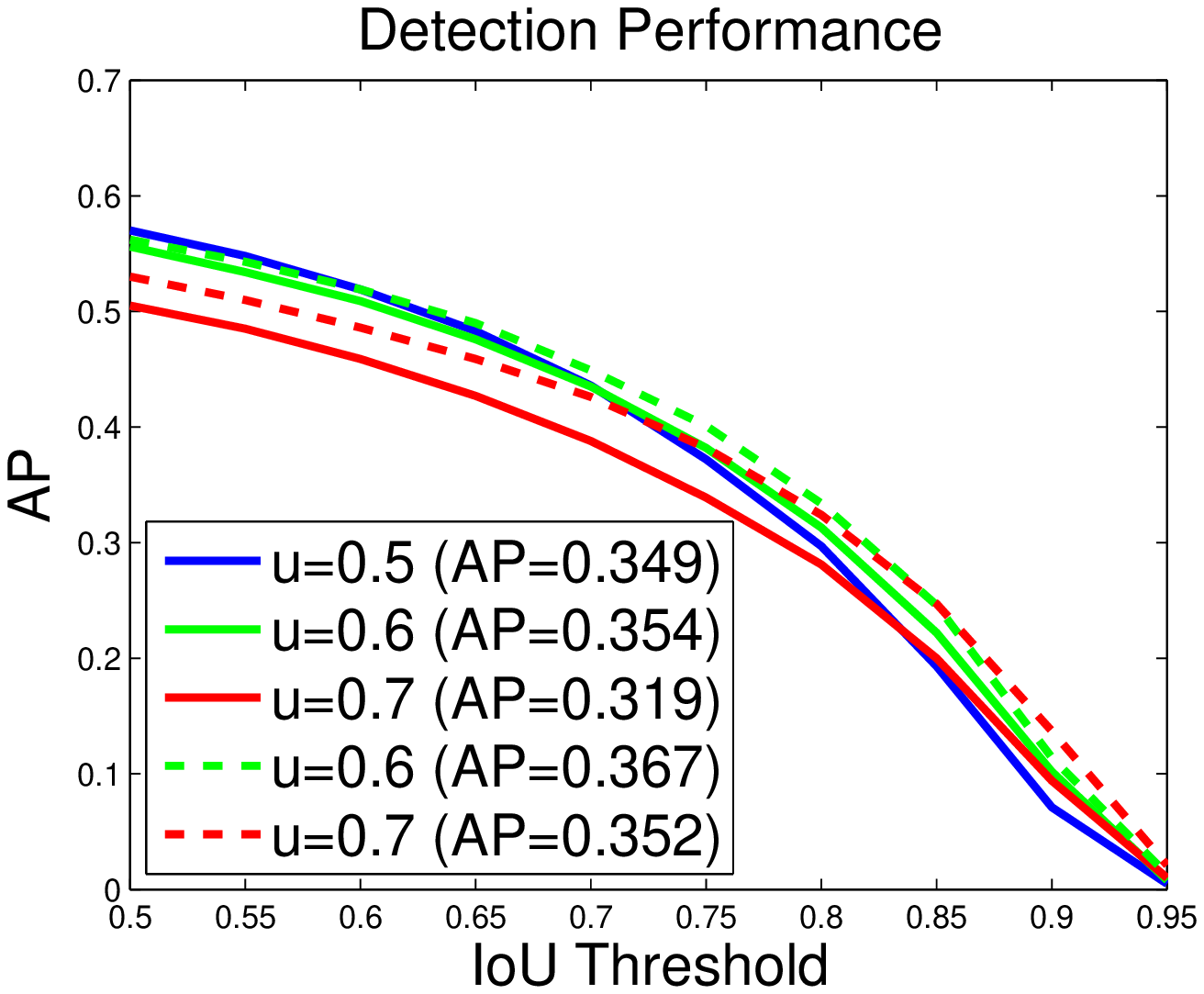,width=4.4cm,height=3.3cm}}{(a)}
\end{minipage}
\hfill
\begin{minipage}[b]{.48\linewidth}
\centering
\centerline{\epsfig{figure=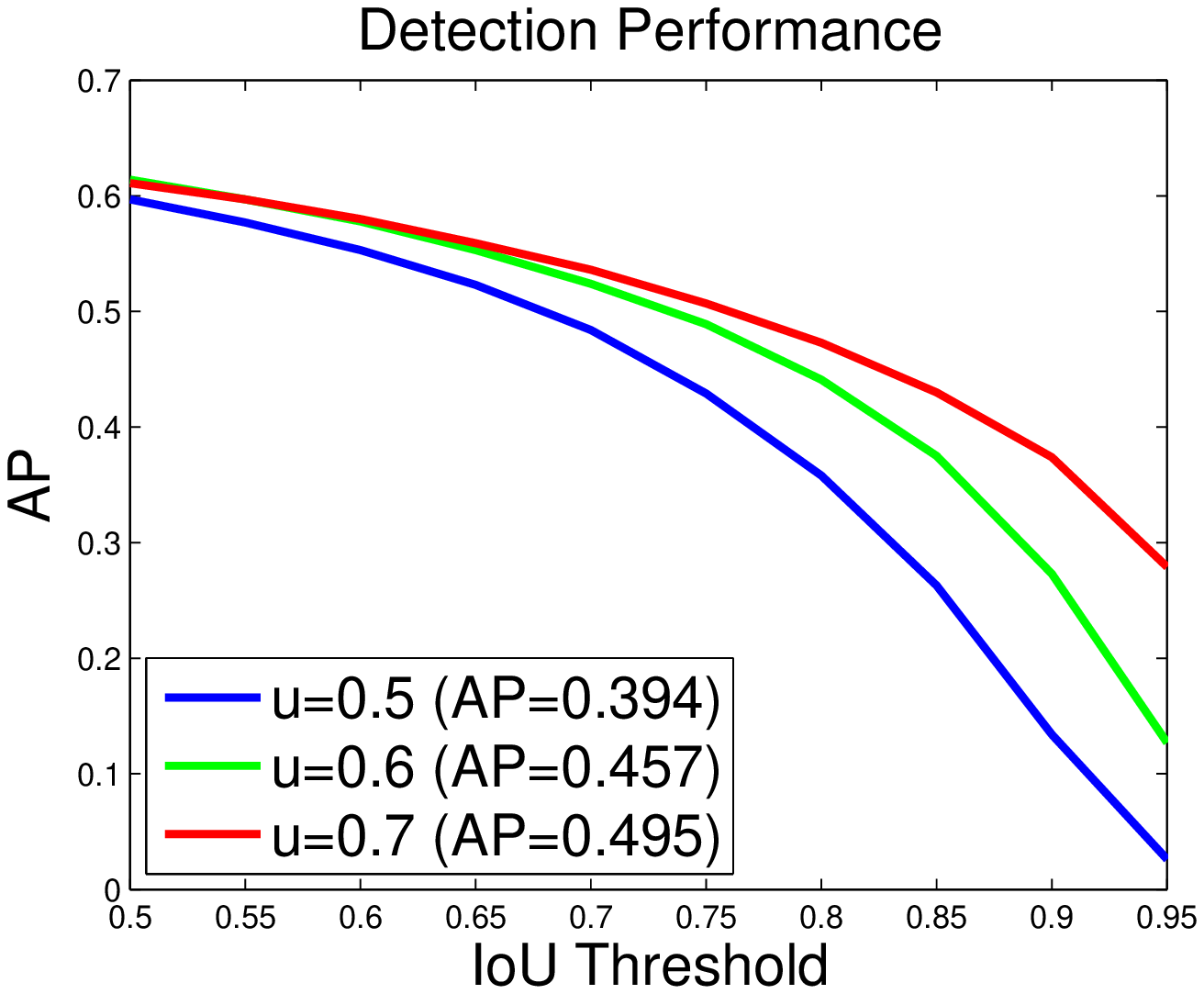,width=4.4cm,height=3.3cm}}{(b)}
\end{minipage}
\caption{(a) detection performance of individually trained detectors, with
their own proposals (solid curves) or Cascade R-CNN stage proposals
(dashed curves). (b) results of adding ground truth to the proposal set.}
\label{fig:hypotheses mismatch}\vspace{-2mm}
\end{figure}

An initial set of experiments was designed to evaluate the impact of
the mistmatch between proposal and detector quality on detection
performance. Figure \ref{fig:hypotheses mismatch} (a) shows the AP
curves of three
individually trained detectors of increasing IoU threshold in
$U=\{0.5,0.6,0.7\}$. The detector of $u=0.5$ outperforms the detector
of $u=0.6$ at low IoU levels, but underperforms it at higher levels.
However, the detector of $u=0.7$ underperforms the other two.
To understand why this happens, we changed the quality of the proposals
at inference. Figure \ref{fig:hypotheses mismatch} (b)
shows the results obtained when ground truth bounding boxes were added to
the set of proposals. While all detectors improved, the detector of $u=0.7$
had the largest gains, and the best performance for almost all
IoU levels. These results suggest two conclusions. First, the
commonly used $u=0.5$ threshold is not effective
for precise detection, simply more robust to low quality
proposals. Second, precise detection requires hypotheses that match
the detector quality.

\begin{figure}[!t]
\begin{minipage}[b]{.3\linewidth}
\centering
\centerline{\epsfig{figure=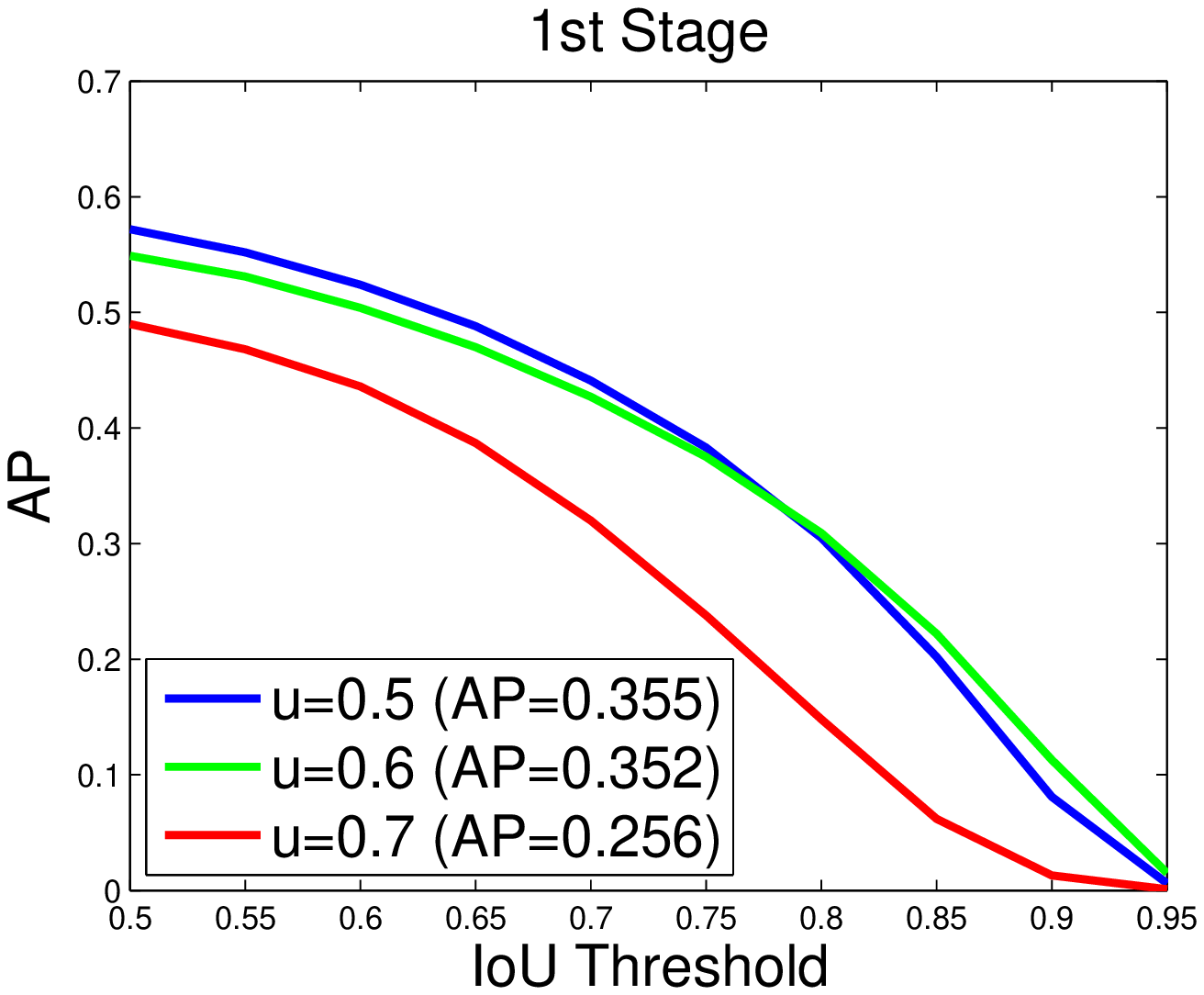,width=3.2cm,height=2.4cm}}
\end{minipage}
\hfill
\begin{minipage}[b]{.3\linewidth}
\centering
\centerline{\epsfig{figure=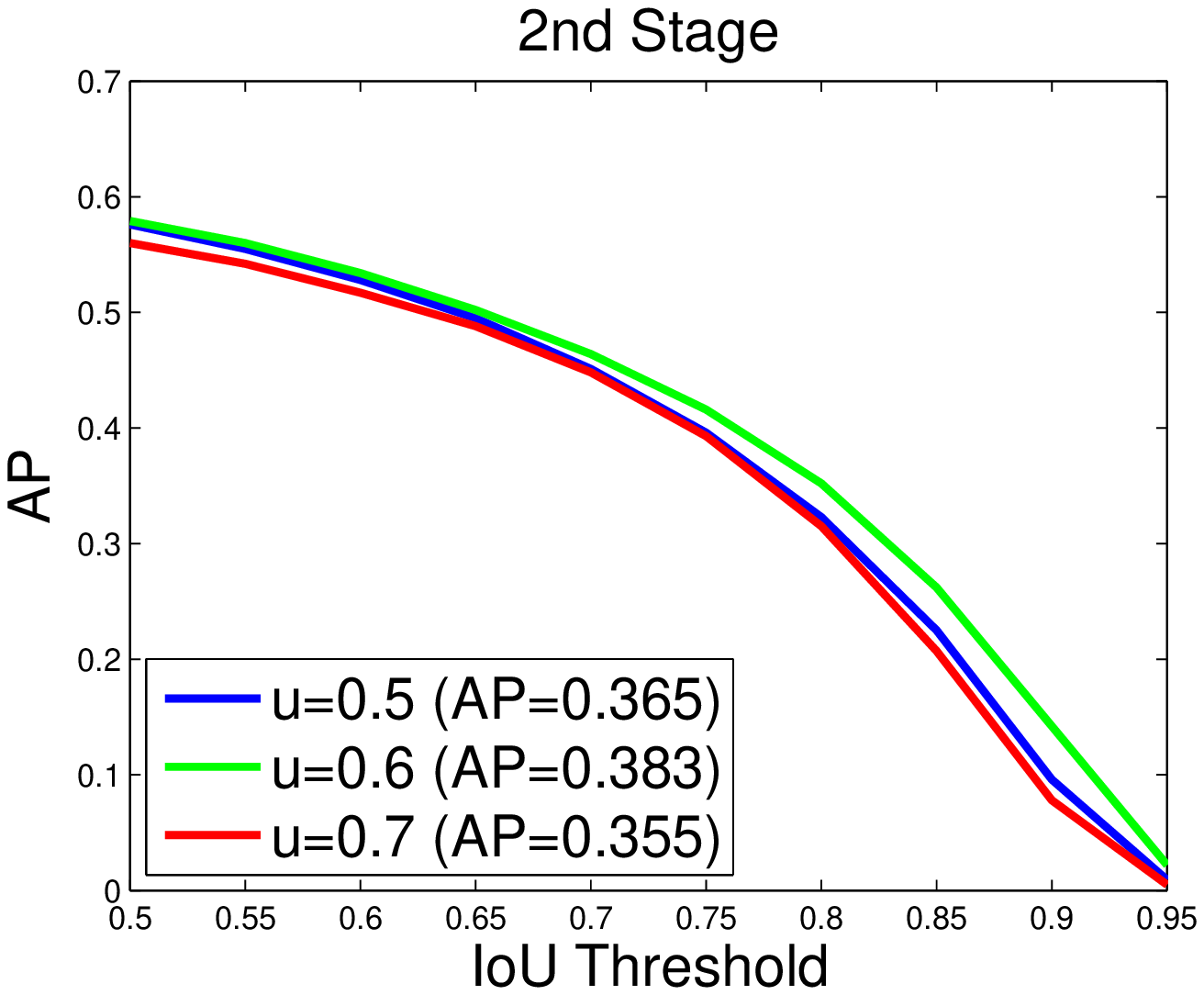,width=3.2cm,height=2.4cm}}
\end{minipage}
\hfill
\begin{minipage}[b]{.3\linewidth}
\centering
\centerline{\epsfig{figure=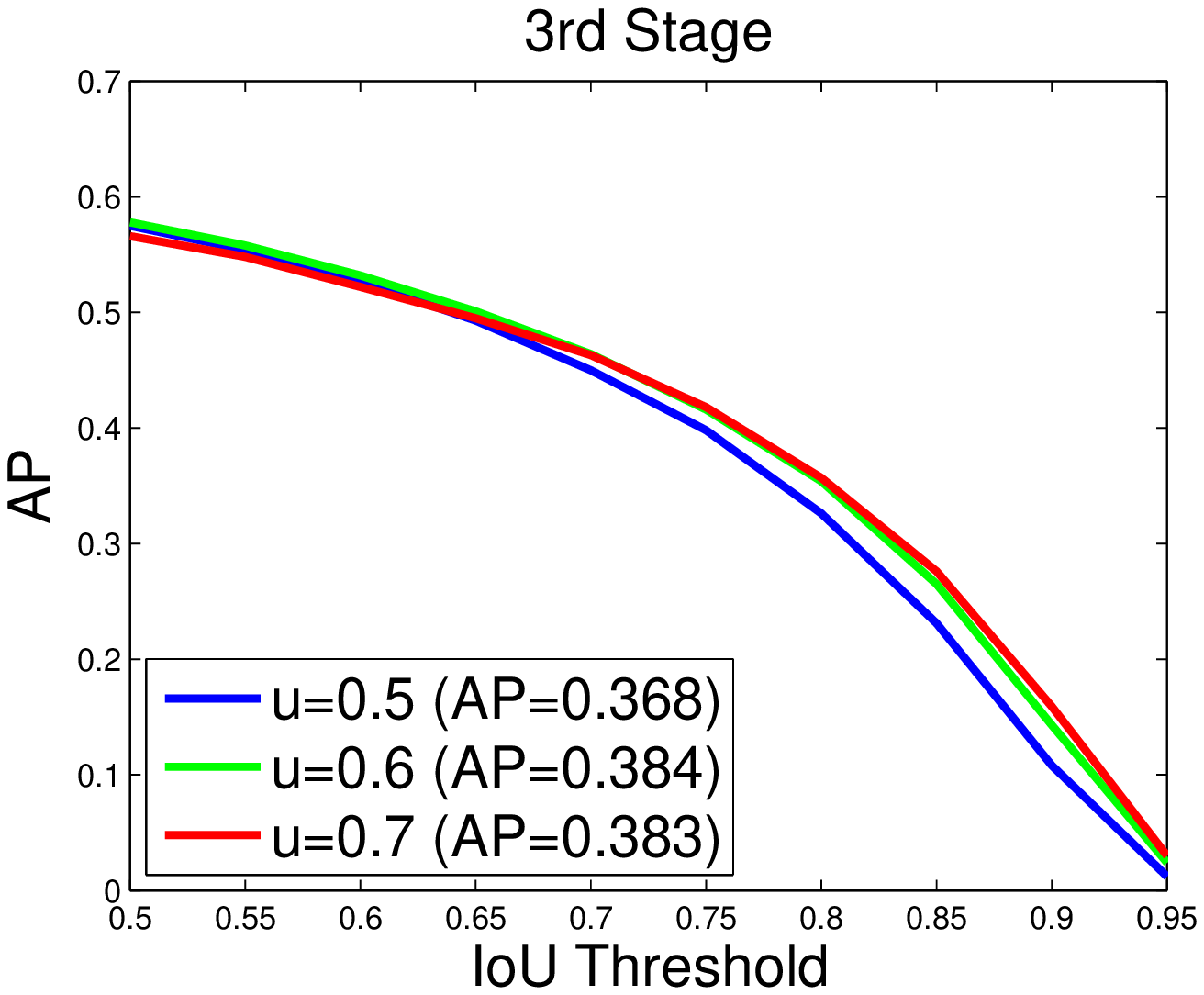,width=3.2cm,height=2.4cm}}
\end{minipage}
\caption{Detection performance of all Cascade R-CNN detectors at all cascade
stages.}
\label{fig:cascade stage}\vspace{-3mm}
\end{figure}

Next, the original proposals were replaced by the
Cascade R-CNN proposals of higher quality ($u=0.6$ and $u=0.7$ used the 2nd
and 3rd stage proposals, respectively). Figure
\ref{fig:hypotheses mismatch} (a) suggests that the performance of the two
detectors is significantly improved when the quality of the test proposals
matches the detector quality.
Testing Cascade R-CNN detectors of different qualities at all cascade stages
produced similar observations. Figure \ref{fig:cascade stage} shows that
each detector was improved by the use of more precise hypotheses, with
higher quality detectors exhibiting larger gains. For example, the detector
of $u=0.7$ performed poorly for the low quality proposals of the 1st stage,
but much better for the more precise hypotheses available at the deeper
cascade stages. The jointly trained detectors of
Fig. \ref{fig:cascade stage} also outperformed the individually
trained detectors of Fig. \ref{fig:hypotheses mismatch} (a), even
when the same proposals were used. This indicates that the detectors are
better trained within the Cascade R-CNN architecture.

\subsection{Comparison with \textit{Iterative BBox} and \textit{Integral Loss}}

In this section, we compare the Cascade R-CNN to
the \textit{iterative BBox} and \textit{integral loss} detectors.
\textit{Iterative BBox} was implemented by applying the detection head of FPN+ baseline
iteratively at inference, three times. The \textit{integral loss} detector was implemented
with three classification heads, using $U=\{0.5,0.6,0.7\}$.

\begin{figure}[!t]
\begin{minipage}[b]{.48\linewidth}
\centering
\centerline{\epsfig{figure=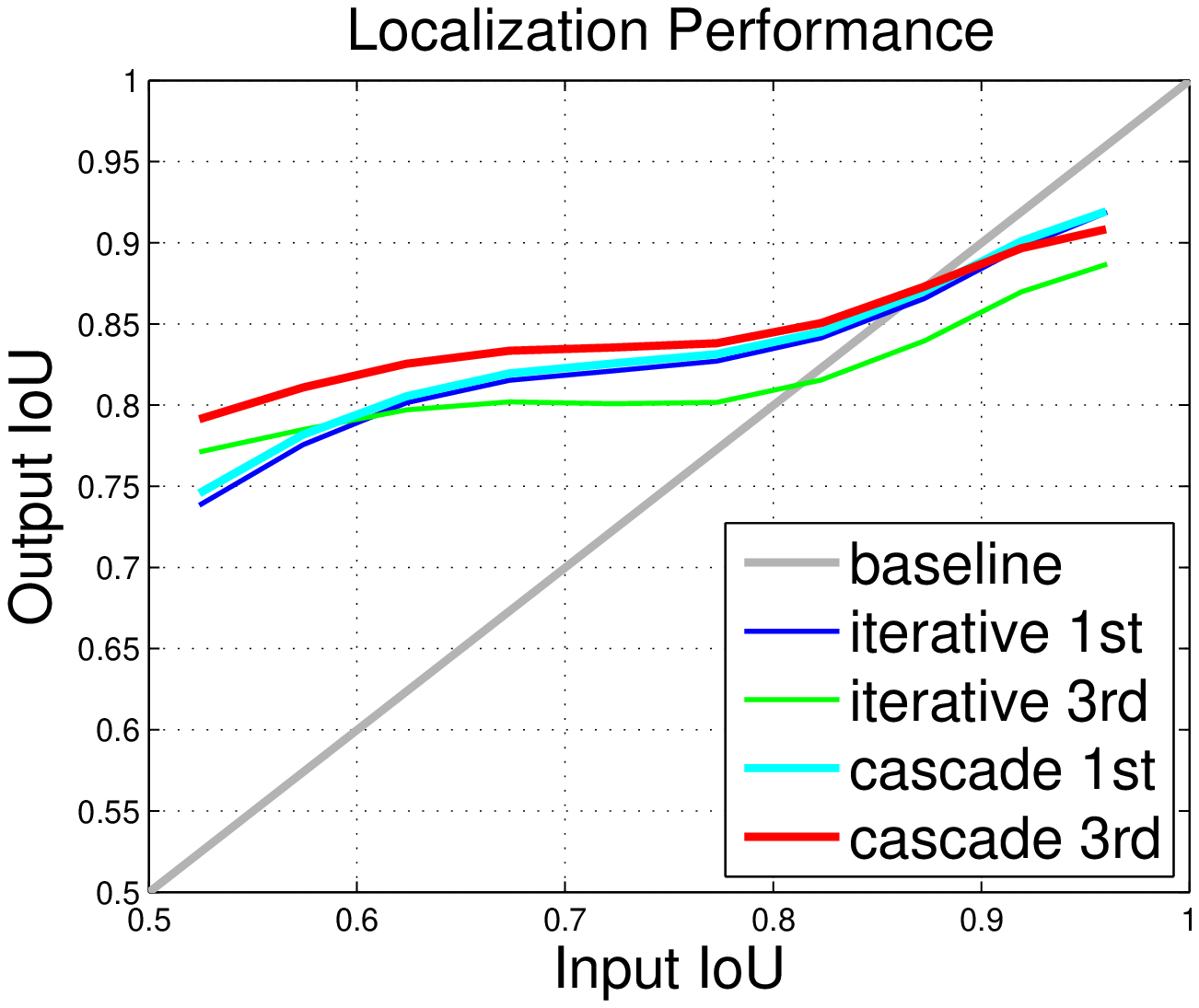,width=4.4cm,height=3.3cm}}{(a)}
\end{minipage}
\hfill
\begin{minipage}[b]{.48\linewidth}
\centering
\centerline{\epsfig{figure=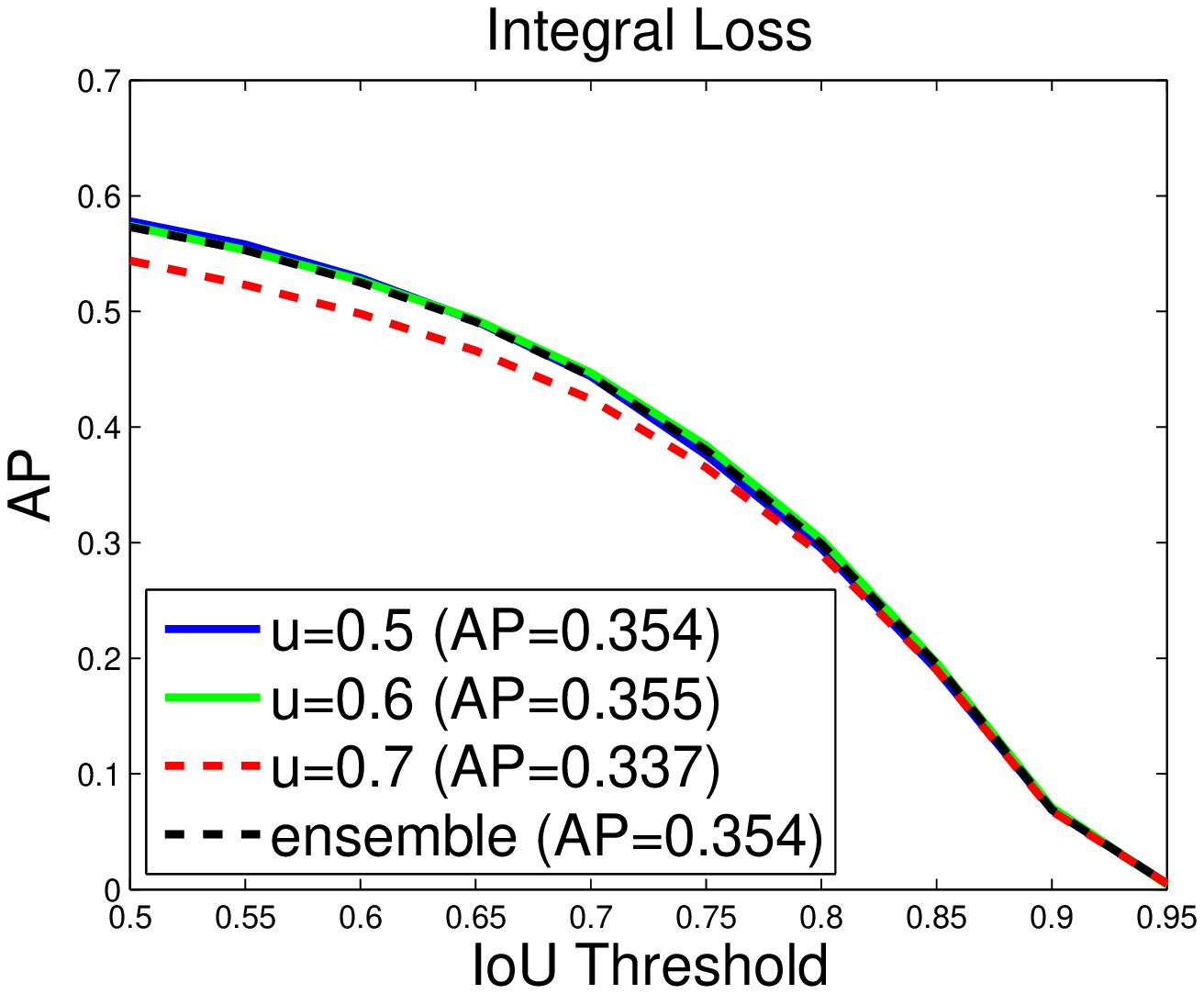,width=4.4cm,height=3.3cm}}{(b)}
\end{minipage}
\caption{(a) localization performance of \textit{iterative BBox} and Cascade
R-CNN regressors. (b) detection performance of the individual
classifiers of the \textit{integral loss} detector.}
\label{fig:localization and integral}\vspace{-2mm}
\end{figure}

\vspace{0.2cm}
\noindent{\bf Localization:}
The localization performances of cascade regression
and \textit{iterative BBox} are compared in
Fig. \ref{fig:localization and integral} (a). The use of a single
regressor degrades localization for hypotheses of high IoU. This effect
accumulates when the regressor is applied iteratively, as
in \textit{iterative BBox}, and performance actually drops with
iteration number. Note the very poor performance of \textit{iterative BBox}
after 3 iterations. On the contrary, the cascade regressor has better
performance at later stages, outperforming \textit{iterative BBox} at almost
all IoU levels. Note that, although cascade regression can slightly
degrade high input IoUs, e.g. IoU$>$0.9, this decrease is
negligible because, as shown in Fig. \ref{fig:hist}, the number of
hypotheses with such high IoUs is extremely small.

\vspace{0.2cm}
\noindent{\bf Integral Loss:} Figure \ref{fig:localization and integral} (b)
summarizes the detection performances of all classifiers
of the \textit{integral loss} detector, sharing a single regressor.
The classifier of $u=0.6$ is the best at all IoU levels, with $u=0.7$
producing the worst results. The ensemble of all classifiers shows no
visible gain.

Table \ref{tab:comparison} shows that both
\textit{iterative BBox} and \textit{integral loss} marginally
improve on the baseline detector, and are not effective for
high quality detection. On the other hand, the Cascade R-CNN
achieves the best performance at all IoU levels. As expected, the gains are
mild for low IoUs, e.g. 0.8 for AP$_{50}$, but significant for the
higher ones, e.g. 6.1 for AP$_{80}$ and 8.7 for AP$_{90}$. Note that
high quality object detection was rarely explored before
this work. These experiments show that 1) it has more room for improvement than
low quality detection, which focuses on AP$_{50}$, and 2) the overall AP can be
significantly improved if it is effectively addressed.

\begin{table}[t]
\tablestyle{1.8pt}{1.2}
\begin{tabular}{l|x{22}|x{22}x{22}x{22}x{22}x{22}}
& AP & AP$_{50}$ & AP$_{60}$ &AP$_{70}$ &AP$_{80}$ &AP$_{90}$\\ [.1em]
\shline
FPN+ baseline &34.9 &57.0  &51.9 &43.6 &29.7  &7.1\\
\textit{Iterative BBox} &35.4 &57.2  &52.1 &44.2 &30.4  &8.1\\
\textit{Integral Loss} &35.4 &57.3  &52.5 &44.4 &29.9  &6.9\\\hline
Cascade R-CNN &\bd{38.9} &\bd{57.8}  &\bd{53.4} &\bd{46.9} &\bd{35.8}  &\bd{15.8}\\
\end{tabular}\vspace{2mm}
\caption{Comparison of the Cascade R-CNN with \textit{iterative BBox} and
\textit{integral loss} detectors.}
\label{tab:comparison}\vspace{-3mm}
\end{table}

\subsection{Ablation Experiments}

A few ablation experiments were run to enable a better understanding of the
Cascade R-CNN.

\vspace{0.2cm}
\noindent{\bf Stage-wise Comparison:}
Table \ref{tab:stage performance} summarizes stagewise performance.
Note that the first stage already outperforms the baseline detector,
due to the benefits of multi-stage multi-task learning. Since deeper cascade
stages prefer higher quality localization, they encourage the learning of
features conducive to it. This benefits the earlier cascade stages, due to
the feature sharing by the backbone network. The second stage improves
performance substantially, and the third is equivalent to the second.
This differs from the \textit{integral loss} detector, where the higher
IoU classifier is relatively weak. While the former (later) stage is better at low (high) IoU metrics, the ensemble of all classifiers is the best overall.

\vspace{0.2cm}
\noindent{\bf IoU Thresholds:}
A Cascade R-CNN was trained using IoU threshold $u=0.5$
for all heads. In this case, the stages differ only in the hypotheses
at their input. Each stage is trained with the corresponding hypotheses, i.e.
accounting for the distribution changes of Fig. \ref{fig:distribution}.
The first row of Table \ref{tab:ablation} shows that this cascade improves
on the baseline detector. This supports the claim that stages should be
optimized for the corresponding sample distributions.
The second row shows performance that improves further when
the threshold $u$ increases across stages.
As discussed in Section \ref{subsec:cascade}, the detector
becomes more selective against close false positives and
{\it specialized\/} to the more precise hypotheses.

\begin{table}[t]
\tablestyle{1.8pt}{1.2}
\begin{tabular}{c|x{22}|x{22}x{22}x{22}x{22}x{22}}
test stage & AP & AP$_{50}$ & AP$_{60}$ &AP$_{70}$ &AP$_{80}$ &AP$_{90}$\\ [.1em]\shline
1 &35.5 &57.2  &52.4 &44.1 &30.5  &8.1\\
2 &38.3 &57.9  &53.4 &46.4 &35.2  &14.2\\
3 &38.3 &56.6  &52.2 &46.3 &35.7  &\bd{15.9}\\
$\overline{1\sim{2}}$ &38.5 &\bd{58.2}  &\bd{53.8} &46.7 &35.0  &14.0\\
$\overline{1\sim{3}}$ &\bd{38.9} &57.8  &53.4 &\bd{46.9} &\bd{35.8}  &15.8\\\hline
FPN+ baseline &34.9 &57.0  &51.9 &43.6 &29.7  &7.1\\
\end{tabular}\vspace{2mm}
\caption{Stagewise performance of the Cascade R-CNN. $\overline{1\sim{3}}$
indicates an ensemble result, obtained by averaging the three classifier
probabilities for 3rd stage proposals.}
\label{tab:stage performance}\vspace{-3mm}
\end{table}

\begin{table}[t]
\tablestyle{1.8pt}{1.2}
\begin{tabular}{x{26}x{28}x{26}|x{22}|x{22}x{22}x{22}x{22}x{22}}
\multirow{2}{*}{IoU$\uparrow$} &\textit{update} &\textit{stage} &\multirow{2}{*}{AP} &\multirow{2}{*}{AP$_{50}$} &\multirow{2}{*}{AP$_{60}$} &\multirow{2}{*}{AP$_{70}$} &\multirow{2}{*}{AP$_{80}$} &\multirow{2}{*}{AP$_{90}$}\\
&\textit{statistics} &\textit{loss} & & & & & &\\ [.1em]\shline
& &\textit{decay}&36.8 &57.8  &52.9 &45.4 &32.0  &10.7\\
\cmark & &\textit{decay} &38.5 &58.4  &54.1 &47.1 &35.0  &13.1\\
&\cmark &\textit{decay} &37.5 &57.8  &53.1 &45.5 &33.3  &13.1\\
\cmark &\cmark &\textit{decay} &38.9 &57.8  &53.4 &46.9 &35.8 &15.8\\
\cmark &\cmark &\textit{avg} &38.9 &57.5  &53.4 &46.9 &35.8  &16.2\\
\end{tabular}\vspace{2mm}
\caption{Ablation experiments. ``IoU$\uparrow$'' indicates increasing IoU
thresholds, ``\textit{update statistics}'' updating regression
statistics, and ``\textit{stage loss}'' weighting of stage losses.}
\label{tab:ablation}\vspace{-3mm}
\end{table}

\vspace{0.2cm}
\noindent{\bf Regression Statistics:}
In Section \ref{subsubsec:bbox}, we saw that the distance vector $\Delta$ is
normalized by the regression statistics (mean and variance), as
in (\ref{eq:norm}). In the Cascade R-CNN, these statistics are
updated stage by stage, as illustrated in
Fig. \ref{fig:distribution}. Updating the statistics of (\ref{eq:norm}) in deeper stages helps the effective multi-task learning of classification and regression. Empirically, the learning is not very sensitive to the exact values of these
statistics. For simplicity, we set $\mu=0$ for all stages,
$\Sigma=(\sigma_x,\sigma_y,\sigma_w,\sigma_h)=(0.1,0.1,0.2,0.2)$ for
the first stage, $\Sigma/2$ for the second, and $\Sigma/3$ for the third,
in all of our experiments. The third and fourth row of
Table \ref{tab:ablation} show that this is beneficial, when compared
to using the statistics of the first stage in all stages (the first and second row).

\vspace{0.2cm}
\noindent{\bf Stage Losses:}
The Cascade R-CNN has multiple detection heads, each with its own loss.
We have explored two schemes to combine these losses:
\textit{decay} and \textit{avg}. In \textit{avg}, the loss of stage $t$
receives a weight $w_t=1/T$, where $T$ is the number of stages.
In \textit{decay}, the weight is $w_t=1/2^{t-1}$. For both schemes, the
learning rate of the head parameters of stage $t$ is rescaled by $1/w_t$,
to ensure that these are sufficiently trained. No rescaling
is needed for the backbone network parameters, since they receive gradients
from all stages. Table \ref{tab:ablation} shows that 1) \textit{avg}
has somewhat better performance for high quality metrics, but worse for
low quality ones, and 2) the two methods have similar overall AP.
The \textit{decay} scheme is used in the remainder of the paper.

\vspace{0.2cm}
\noindent{\bf Number of Stages:}
Table \ref{tab:cascade stage} summarizes the impact of the number of stages
in the Cascade R-CNN performance. Adding a second
stage significantly improves the baseline detector.
Three detection stages still produce non-trivial improvement, but the
addition of a $4^{th}$ stage ($u=0.75$) has a slight performance
decrease. Note, however, that while the overall AP degrades,
the four-stage cascade has the best performance at high IoU levels. The
three-stage cascade achieves the best trade-off between cost and
AP performance, and is used in the remaining experiments.

\begin{table}[t]
\tablestyle{1.8pt}{1.2}
\begin{tabular}{c|c|x{22}|x{22}x{22}x{22}x{22}x{22}}
\# stages &test stage & AP & AP$_{50}$ & AP$_{60}$ &AP$_{70}$ &AP$_{80}$ &AP$_{90}$\\ [.1em]\shline
1 &1 &34.9 &57.0  &51.9 &43.6 &29.7  &7.1\\
2 &$\overline{1\sim{2}}$ &38.2 &\bd{58.0}  &\bd{53.6} &46.7 &34.6  &13.6\\
3 &$\overline{1\sim{3}}$ &\bd{38.9} &57.8  &53.4 &\bd{46.9} &35.8  &15.8\\
4 &$\overline{1\sim{3}}$ &\bd{38.9} &57.4  &53.2 &46.8 &\bd{36.0}  &16.0\\
4 &$\overline{1\sim{4}}$ &38.6 &57.2  &52.8 &46.2 &35.5  &\bd{16.3}\\
\end{tabular}\vspace{2mm}
\caption{The impact of the number of stages in Cascade R-CNN.}
\label{tab:cascade stage}\vspace{-3mm}
\end{table}

\begin{table*}[t]
\tablestyle{3.5pt}{1.1}
\begin{tabular}{l|l|x{22}x{22}x{22}|x{22}x{22}x{22}}
&backbone &AP &AP$_{50}$ &AP$_{75}$  &AP$_{S}$ &AP$_{M}$ &AP$_{L}$\\\shline
YOLOv2 \cite{DBLP:conf/cvpr/RedmonDGF16}       &DarkNet-19 &21.6 &44.0 &19.2  &5.0 &22.4 &35.5\\
SSD513 \cite{DBLP:conf/eccv/LiuAESRFB16}$^{\ast}$    &ResNet-101 &31.2 &50.4 &33.3  &10.2 &34.5 &49.8\\
RetinaNet \cite{lin2017focal}$^{\ast}$ &ResNet-101  &39.1 &59.1 &42.3  &21.8 &42.7 &50.2\\
CornerNet \cite{DBLP:conf/eccv/LawD18}$^{\ast\star}$ &Hourglass-104  &42.1 &57.8 &45.3  &20.8 &44.8 &56.7\\\hline
Faster R-CNN+++ \cite{DBLP:conf/cvpr/HeZRS16}$^{\ast\star}$      &ResNet-101 &34.9 &55.7 &37.4  &15.6 &38.7 &50.9\\
Faster R-CNN w FPN \cite{lin2017feature}   &ResNet-101 &36.2 &59.1 &39.0  &18.2 &39.0 &48.2\\
Faster R-CNN w FPN+ (ours) &ResNet-101 &38.8 &61.1 &41.9  &21.3 &41.8 &49.8\\
G-RMI \cite{DBLP:journals/corr/HuangRSZKFFWSG016}$^{\ast\star}$ &Inception-ResNet-v2 &41.6 &62.3 &45.6  &24.0 &43.9 &55.2\\
Deformable R-FCN \cite{dai2017deformable}$^{\ast\star}$   &Aligned-Inception-ResNet &37.5 &58.0 &40.8  &19.4 &40.1 &52.5\\
Mask R-CNN \cite{he2017mask}       &ResNet-101 &38.2 &60.3 &41.7  &20.1 &41.1 &50.2\\
RelationNet \cite{hu2018relation} &ResNet-101  &39.0 &58.6 &42.9  &- &- &-\\
DetNet \cite{DBLP:conf/eccv/LiPYZDS18} &DetNet-59   &40.3 &62.1 &43.8  &23.6 &42.6 &50.0\\
SNIP \cite{singh2018analysis}$^{\ast\star}$ &DPN-98  &45.7 &67.3 &51.1  &29.3 &48.8 &57.1\\\hline
AttractioNet \cite{DBLP:conf/bmvc/GidarisK16}$^{\star}$ &VGG16+Wide ResNet &35.7 &53.4 &39.3  &15.6 &38.0 &52.7\\
\bd{Cascade R-CNN} &ResNet-101 &42.8 &62.1 &46.3  &23.7 &45.5 &55.2\\
\bd{Cascade R-CNN}$^{\ast\star}$ &ResNeXt-152 &50.9 &69.0 &55.8  &33.4 &53.5  &63.3\\\hline
\end{tabular}
\vspace{0.1cm}
\caption{Performance of state-of-the-art \emph{single-model} detectors on
COCO \texttt{test-dev}. Entries denoted by $^{\ast}$ and $^{\star}$ use
enhancements at training and inference, respectively.}
\label{tab:state-of-the-art}\vspace{-2mm}
\end{table*}

\begin{table*}[t]
\tablestyle{1.8pt}{1.2}
\begin{tabular}{c|c|c|x{22}|x{22}|x{22}|x{18}x{18}x{18}x{18}x{18}x{18}|x{18}x{18}x{18}x{18}x{18}x{18}}
& \multirow{2}{*}{backbone} & \multirow{2}{*}{cascade} & train & test & model &\multicolumn{6}{c|}{\texttt{val} (5k)} &\multicolumn{6}{c}{\texttt{test-dev} (20k)}\\\cline{7-18}
& & &speed &speed &size &AP & AP$_{50}$ & AP$_{75}$ &AP$_{S}$ &AP$_{M}$ &AP$_{L}$ & AP & AP$_{50}$ & AP$_{75}$ &AP$_{S}$ &AP$_{M}$ &AP$_{L}$\\ [.1em]
\shline
\multirow{2}{*}{Faster R-CNN} &\multirow{2}{*}{VGG} & \xmark &0.12s & 0.075s &278M &23.6 &43.9  &23.0 &8.0 &26.2  &35.5 &23.5 &43.9  &22.6 &8.1 &25.1  &34.7\\
& & \cmark &0.14s & 0.115s &704M &27.0 &44.2  &27.7 &8.6 &29.1  &42.2 &26.9 &44.3  &27.8 &8.3 &28.2  &41.1\\\hline
\multirow{2}{*}{R-FCN} &\multirow{2}{*}{ResNet-50} & \xmark &0.19s & 0.07s &133M &27.0 &48.7  &26.9 &9.8 &30.9  &40.3 &27.1 &49.0  &26.9 &10.4 &29.7  &39.2\\
& & \cmark &0.24s & 0.075s &184M &31.1 &49.8  &32.8 &10.4 &34.4  &48.5 &30.9 &49.9  &32.6 &10.5 &33.1  &46.9\\\hline
\multirow{2}{*}{R-FCN} &\multirow{2}{*}{ResNet-101} & \xmark &0.23s & 0.075s &206M &30.3 &52.2  &30.8 &12.0 &34.7  &44.3 &30.5 &52.9  &31.2 &12.0 &33.9  &43.8\\
& & \cmark &0.29s & 0.083s &256M &33.3 &52.0  &35.2 &11.8 &37.2  &51.1 &33.3 &52.6  &35.2 &12.1 &36.2  &49.3\\\hline
\multirow{2}{*}{FPN+} &\multirow{2}{*}{ResNet-50} & \xmark &0.30s & 0.095s &165M &36.5 &58.6  &39.2 &20.8 &40.0  &47.8 &36.5 &59.0  &39.2 &20.3 &38.8  &46.4\\
& & \cmark &0.33s & 0.115s &272M &40.3 &59.4  &43.7 &22.9 &43.7  &54.1 &40.6 &59.9  &44.0 &22.6 &42.7  &52.1\\\hline
\multirow{2}{*}{FPN+} &\multirow{2}{*}{ResNet-101} & \xmark &0.38s & 0.115s &238M &38.5 &60.6  &41.7 &22.1 &41.9  &51.1 &38.8 &61.1 &41.9  &21.3 &41.8 &49.8\\
& & \cmark &0.41s & 0.14s &345M &42.7 &61.6  &46.6 &23.8 &46.2  &57.4 &42.8 &62.1 &46.3  &23.7 &45.5 &55.2\\\hline
\end{tabular}\vspace{2mm}
\caption{Performance of Cascade R-CNN implementations with multiple
detectors. All speeds are reported per image on a single Titan Xp GPU.}
\label{tab:generalization}\vspace{-3mm}
\end{table*}

\subsection{Comparison with the state-of-the-art}
\label{subsec:state-of-the-art}

An implementation of the Cascade R-CNN, based on the FPN+ detector and the ResNet-101 backbone, is compared to state-of-the-art \emph{single-model}
detectors in Table \ref{tab:state-of-the-art}\footnote{Some
detectors are omitted in this comparison because their \emph{single-model}
results on COCO \texttt{test-dev} are not publicly available.}. The
settings are those of Section \ref{subsubsec:baseline}, but training
used 280k iterations, with learning rate decreased at 160k and 240k
iterations. The number of RoIs was also increased to 512. The top
of the table reports to one-stage detectors, the middle to two-stage, and the
bottom to multi-stage (3-stages+RPN for the Cascade R-CNN). Note that
{\it all} the compared state-of-the-art detectors are trained
with $u=0.5$.

An initial observation is that our FPN+ implementation is
better than the original FPN \cite{lin2017feature}, providing a very strong
baseline. Nevertheless, the extension from FPN+ to Cascade R-CNN improved
performance by $\sim$4 points. In fact, the vanilla Cascade R-CNN, without
any bells and whistles, outperformed almost all \emph{single-model} detectors
under all evaluation metrics. This includes the COCO challenge 2016 winner
G-RMI \cite{DBLP:journals/corr/HuangRSZKFFWSG016}, the recent
Deformable R-FCN \cite{dai2017deformable}, RetinaNet \cite{lin2017focal},
Mask R-CNN \cite{he2017mask}, RelationNet\cite{hu2018relation},
DetNet\cite{DBLP:conf/eccv/LiPYZDS18}, CornerNet \cite{DBLP:conf/eccv/LawD18},
etc. Note some of these methods leverage several training or inference enhancements,
e.g. multi-scale, soft NMS \cite{DBLP:conf/iccv/BodlaSCD17}, etc,
making the comparison very unfair. Finally, compared to the previously best
multi-stage detector on COCO, AttractioNet \cite{DBLP:conf/bmvc/GidarisK16}, the vanilla Cascade R-CNN has a gain of 7.1
points.

The only detector that outperforms the Cascade R-CNN in
Table \ref{tab:state-of-the-art} is SNIP \cite{singh2018analysis},
which uses multi-scale training and inference, a larger input size,
a stronger backbone, Soft NMS, and some other enhancements. For a more fair
comparison, we implemented the Cascade R-CNN with multi-scale
training/inference, a stronger backbone (ResNeXt-152 \cite{DBLP:conf/cvpr/XieGDTH17}), mask supervision, etc.
This enhanced Cascade R-CNN surpassed SNIP by 5.2 points. It
also outperforms the \emph{single-model} MegDet detector (50.6 mAP),
which won the COCO challenge in 2017 and uses many other
enhancements \cite{peng2018megdet}. The Cascade R-CNN is conceptually
straightforward, simple to implement, and can be combined, in a plug and play
manner, with many detector architectures.

\subsection{Generalization Capacity}

To more thoroughly test this claim, a three-stage Cascade R-CNN
was implemented with three baseline detectors: Faster R-CNN,
R-FCN, and FPN+. All settings are as discussed above,
with the variations discussed in Section \ref{subsec:state-of-the-art} for
the FPN+ detector. Table \ref{tab:generalization} presents a comparison
of the AP performance of the three detectors.

\vspace{0.2cm}
\noindent{\bf Detection Performance:}
Again, our implementations are better than the original
detectors \cite{DBLP:conf/nips/RenHGS15,DBLP:conf/nips/DaiLHS16,
lin2017feature}. Still, the Cascade R-CNN improves on all baselines
by 2$\sim$4 points, independently of their strength.
Similar gains are observed for \texttt{val} and \texttt{test-dev}. These
results show that the Cascade R-CNN is widely applicable across
detector architectures.

\vspace{0.2cm}
\noindent{\bf Parameters and Timing:}
The number of Cascade R-CNN parameters increases with the number of
stages. The increase is linear and proportional
to the parameter cardinality of the baseline detector head.
However, because the head has much less computation than the backbone
network, the Cascade R-CNN has
small computational overhead, at both training and testing.
This is shown in Table \ref{tab:generalization}.

\vspace{0.2cm}
\noindent{\bf Codebase and Backbone:}
The Cascade R-CNN of FPN was also reimplemented on the Detectron
codebase \cite{Detectron2018} with various backbone networks.
Table \ref{tab:detectron} summarizes these experiments, showing very
consistent improvements (3$\sim$4 points) across backbones.
The Cascade R-CNN has also been independently reproduced by other research
groups, on PyTorch and TensorFlow. These again
show that the Cascade R-CNN can provide reliable gains
across detector architectures, backbones, codebases, and implementations.

\begin{table}[t]
\tablestyle{1.8pt}{1.2}
\begin{tabular}{c|c|x{18}x{18}x{18}x{18}x{18}x{18}}
backbone &cascade &AP &AP$_{50}$ &AP$_{75}$ &AP$_{S}$ &AP$_{M}$ &AP$_{L}$\\ [.1em]
\shline
\multirow{2}{*}{Fast ResNet-50} & \xmark &36.4 &58.4  &39.3 &20.3 &39.8  &48.1\\
& \cmark &40.5 &58.7  &43.9 &21.5 &43.6  &54.9\\\hline
\multirow{2}{*}{ResNet-50} & \xmark &36.7 &58.4  &39.6 &21.1 &39.8  &48.1\\
& \cmark &40.9 &59.0  &44.6 &22.5 &43.6  &55.3\\\hline
\multirow{2}{*}{ResNet-101} & \xmark &39.4 &61.2  &43.4 &22.6 &42.9  &51.4\\
& \cmark &42.8 &61.4  &46.8 &24.1 &45.8  &57.4\\\hline
\multirow{2}{*}{ResNeXt-101} & \xmark &41.3 &63.7  &44.7 &25.5 &45.3  &52.9\\
& \cmark &44.7 &63.7  &48.8 &26.3 &48.4  &58.6\\\hline
\multirow{2}{*}{ResNet-50-GN} & \xmark &38.4 &59.9  &41.7 &22.2 &41.2  &50.0\\
& \cmark &42.2 &60.6  &45.8 &24.7 &45.2  &55.7\\\hline
\multirow{2}{*}{ResNet-101-GN} & \xmark &39.9 &61.3  &43.3 &23.6 &42.8  &52.3\\
& \cmark &43.8 &62.2  &47.6 &26.2 &47.2  &57.7\\\hline
\end{tabular}\vspace{2mm}
\caption{Performance of various implementations of the Cascade R-CNN with
the FPN detector on Detectron, using the \texttt{1x} schedule.}
\label{tab:detectron}\vspace{-3mm}
\end{table}

\vspace{0.2cm}
\noindent{\bf Fast R-CNN:}
As shown in Fig. \ref{fig:framework} (b), the Cascade R-CNN is not
limited to the standard Faster R-CNN architecture. To test this, we trained the
Cascade R-CNN in the way of the Fast R-CNN, using
pre-collected proposals. The results of Table \ref{tab:detectron}
show that the gains of the Cascade R-CNN hold for frameworks other than
the Faster R-CNN.

\vspace{0.2cm}
\noindent{\bf Group Normalization:}
Group normalization (GN) \cite{DBLP:conf/eccv/WuH18} is a recent
normalization technique, published after the Cascade R-CNN. It addresses
the problem that batch normalization (BN) \cite{DBLP:conf/icml/IoffeS15}
must be frozen for object detector training, due to the inaccurate statistics
that can be derived from small batch sizes \cite{peng2018megdet}. GN,
an alternative to BN that is independent of batch size, has comparable
performance to large-batch synchronized BN. Table \ref{tab:detectron}
shows that the Cascade R-CNN with GN has similar gains to those
obersved for the other architectures. This suggests that the Cascade R-CNN
will continue to be useful even as architectural enhancements continue to
emerge in the literature.

\begin{table}[t]
\tablestyle{1.8pt}{1.2}
\begin{tabular}{c|x{22}x{22}x{22}x{22}|x{22}x{22}x{22}x{22}}
stage &AP$^{100}$ &AP$^{100}_{s}$ &AP$^{100}_{m}$ &AP$^{100}_{l}$ &AP$^{1k}$ &AP$^{1k}_{s}$ &AP$^{1k}_{m}$ &AP$^{1k}_{l}$\\ [.1em]\shline
FPN &47.8 &32.2  &54.9 &65.2 &59.1  &48.0 &66.3  &68.4\\\hline
1 &46.8 &31.0  &53.8 &64.8 &58.7  &47.6 &65.9  &68.2\\
2 &55.3 &35.1  &61.1 &82.5 &70.7  &55.2 &77.7  &88.1\\
3 &56.5 &36.1  &62.4 &84.1 &71.4  &55.5 &78.1  &89.8\\
\end{tabular}\vspace{2mm}
\caption{Proposal recall of Cascade R-CNN stages.}
\label{tab:proposal}\vspace{-3mm}
\end{table}

\subsection{Proposal Evaluation}

Table \ref{tab:proposal} summarizes the proposal recall performance of a
Cascade R-CNN implemented with the FPN detector and ResNet-50 backbone.
The first Cascade R-CNN stage has proposal recall close to that
of the FPN baseline. The addition of a bounding box regression stage improves
recall significantly, e.g. from 59.1 to 70.7 for AP$^{1k}$
and close to 20 points for AP$^{1k}_{l}$. This shows that the additional
bounding box regression is very effective at improving proposal recall
performance. The addition of a third stage has a smaller but non-negligible
gain. This high
proposal recall performance secures the later high-quality
object detection task.

\subsection{Instance Segmentation by Cascade Mask R-CNN}

Table \ref{tab:mask solutions} summarizes the instance segmentation performance of the
Cascade Mask R-CNN strategies of
Fig. \ref{fig:maskframework}. These experiments, use the Mask R-CNN,
implemented on Detectron with \texttt{1x} schedule as baseline.
All three strategies improve on baseline performance, although with smaller
gains than object detection (see Table \ref{tab:stage performance}),
especially at high quality. For example, the AP$_{90}$
improvement of 8.7 points for object detection falls to
1.8 points, showing that plenty
of room is left for improving high quality instance segmentation. Comparing
strategies, (c) outperforms (b).
This is because (b) trains
the mask head in the first stage but tests after the last stage,
leading to a mask prediction mismatch. This mismatch is reduced by
(c). The addition of a mask branch to each stage by strategy (d) does not
have noticeable benefits over (c), but requires much more computation
and memory. Strategy (b) has the best trade-off between cost and AP
performance, and is used in the remainder of the paper.

To evaluate the instance segmentation robustness of the Cascade Mask R-CNN,
several backbone networks are compared in Table \ref{tab:mask}.
Since this architecture can detect objects, detection results
are also shown. Note that the additional mask supervision makes these
better than those of Table \ref{tab:detectron}.
The gains of the Cascade Mask R-CNN are very consistent for all backbone networks.
Even when the strongest model, ResNeXt-152 \cite{DBLP:conf/cvpr/XieGDTH17},
is used with training data augmentation and \texttt{1.44x} schedule, the
Cascade Mask R-CNN has a gain of 2.9 points for detection and 1.0 point for
instance segmentation. Adding inference enhancements, the gains are
still 2.1 points for detection and 0.8 points for instance segmentation.
This robustness explains why the Cascade R-CNN was widely used in the
COCO challenge 2018, where the task is instance segmentation, not
object detection.

\begin{table}[t]
\tablestyle{1.8pt}{1.2}
\begin{tabular}{l|x{22}|x{22}x{22}x{22}x{22}x{22}}
& AP & AP$_{50}$ & AP$_{60}$ &AP$_{70}$ &AP$_{80}$ &AP$_{90}$\\ [.1em]
\shline
Mask R-CNN baseline &33.9 &55.5  &49.8 &41.4 &28.6  &8.3\\\hline
strategy of Fig. \ref{fig:maskframework} (b) &35.0 &56.3  &50.8 &43.0 &30.0  &9.3\\
strategy of Fig. \ref{fig:maskframework} (c) &35.4 &56.4  &50.9 &43.2 &31.0  &10.1\\
strategy of Fig. \ref{fig:maskframework} (d) &35.5 &56.5  &51.2 &43.4 &30.8  &10.0\\
\end{tabular}\vspace{2mm}
\caption{The instance segmentation comparison among three strategies of the Cascade Mask R-CNN.}
\label{tab:mask solutions}\vspace{-3mm}
\end{table}

\begin{table*}[htp]
\tablestyle{1.8pt}{1.2}
\begin{tabular}{c|c|x{22}x{22}x{22}x{22}x{22}x{22}|x{22}x{22}x{22}x{22}x{22}x{22}}
\multirow{2}{*}{backbone} & \multirow{2}{*}{cascade} &\multicolumn{6}{c|}{Object Detection} &\multicolumn{6}{c}{Instance Segmentation}\\\cline{3-14}
& &AP & AP$_{50}$ & AP$_{75}$ &AP$_{S}$ &AP$_{M}$ &AP$_{L}$ &AP & AP$_{50}$ & AP$_{75}$ &AP$_{S}$ &AP$_{M}$ &AP$_{L}$\\ [.1em]
\shline
\multirow{2}{*}{ResNet-50} & \xmark &37.7 &59.2  &40.9 &21.4 &40.8  &49.8 &33.9  &55.8 &35.8  &14.9 &36.3 &50.9\\
& \cmark &41.3 &59.4 &45.3 &23.2 &43.8  &55.8 &35.4  &56.4 &37.7 &15.9 &37.7  &53.6\\\hline
\multirow{2}{*}{ResNet-101} & \xmark &40.0 &61.8  &43.7 &22.5 &43.4  &52.7 &35.9  &58.3 &38.0  &15.9 &38.9 &53.2\\
& \cmark &43.3 &61.7  &47.2 &24.2 &46.3 &58.2 &37.1  &58.6 &39.8  &16.7 &39.7 &55.7\\\hline
\multirow{2}{*}{ResNet-50-GN} & \xmark &39.2 &60.5  &42.9 &22.9 &42.2  &50.6 &34.9  &57.1 &36.9  &16.0 &37.7 &51.2\\
& \cmark &42.9 &60.7 &46.6 &25.1 &45.9  &56.7 &36.6  &57.7 &39.2 &16.8 &39.3  &54.5\\\hline
\multirow{2}{*}{ResNet-101-GN} & \xmark &41.1 &62.1  &45.1 &23.6 &44.3  &53.1 &36.3  &58.9 &38.5  &16.2 &39.4 &53.6\\
& \cmark &44.8 &62.8  &48.8 &26.4 &48.0 &58.7 &38.0  &59.8 &40.8  &18.1 &40.7 &56.0\\\hline
\multirow{2}{*}{ResNeXt-101} & \xmark &42.1 &64.1  &45.9 &25.6 &45.9  &54.4 &37.3  &60.3 &39.5  &17.8 &40.3 &55.5\\
& \cmark &45.8 &64.1 &50.3  &27.2 &49.5 &60.1 &38.6 &60.6  &41.5  &18.5 &41.3 &57.2 \\\hline
\multirow{2}{*}{ResNeXt-152$^{\ast}$} & \xmark &45.2 &66.9  &49.7 &28.5 &49.4  &56.8 &39.7  &63.5 &42.4  &19.8 &42.9 &57.3\\
& \cmark &48.1 &66.7 &52.6 &29.3 &52.2  &62.1 &40.7 &63.7  &43.8  &19.9 &44.0 &59.1 \\\hline
\multirow{2}{*}{ResNeXt-152$^{\ast\star}$} & \xmark &48.1 &68.3  &52.9 &32.6 &51.8  &61.3 &41.5  &65.1 &44.7  &22.0 &44.8 &59.8\\
& \cmark &50.2 &68.2 &55.0  &33.1 &53.9  &64.2 &42.3  &65.4 &45.8  &21.9 &45.7 &60.9 \\\hline
\end{tabular}\vspace{2mm}
\caption{Performance of the Cascade Mask R-CNN on multiple backbone networks on COCO 2017 \texttt{val}. $^{\ast}$ and $^{\star}$ denotes enhancement techniques at training and inference, respectively, as in \cite{Detectron2018}.}
\label{tab:mask}\vspace{-3mm}
\end{table*}

\subsection{Results on PASCAL VOC}
\label{subsec:voc}

The Cascade R-CNN was further tested on the PASCAL VOC
dataset \cite{DBLP:journals/ijcv/EveringhamGWWZ10}.
Following \cite{DBLP:conf/nips/RenHGS15,DBLP:conf/eccv/LiuAESRFB16}, the
models were trained on VOC2007 and VOC2012 \texttt{trainval} (16,551
images) and tested on VOC2007 \texttt{test} (4,952 images). Two
detector architectures were evaluated: Faster R-CNN (with AlexNet and
VGG-Net backbones) and R-FCN (with ResNet-50 and
ResNet-101). Training details were as discussed in
Section \ref{subsubsec:baseline}, and both AlexNet and VGG-Net were pruned.
More specifically, Faster R-CNN (R-FCN) training started with a learning
rate of 0.001 (0.002), which was reduced by a factor of 10 at 30k (60k) and
stopped at 45k (90k) iterations. Since the standard VOC evaluation
metric (AP at IoU of 0.5) is fairly saturated, and the focus of this work is
high quality detection, the COCO metrics were used
for evaluation\footnote{The PASCAL VOC annotations were transformed to COCO
format, and the COCO toolbox used for evaluation. Results are different
from the standard VOC evaluation.}. Table \ref{tab:voc} summarizes the
performance of all detectors, showing that the Cascade R-CNN significantly
improves the overall AP in all cases. These results are further
evidence for the robustness of the Cascade R-CNN.

\subsection{Additional Results on other Datasets}
\label{subsec:kitti+cityperson}

Beyond generic object detection datasets, the Cascade R-CNN was tested
on some specific object detection tasks, including
KITTI \cite{DBLP:conf/cvpr/GeigerLU12},
CityPerson \cite{DBLP:conf/cvpr/ZhangBS17} and
WiderFace \cite{DBLP:conf/cvpr/YangLLT16}. The
MS-CNN \cite{DBLP:conf/eccv/CaiFFV16}, a detector of strong performance on
these tasks, was used as baseline for all of them.

\vspace{0.2cm}
\noindent{\bf KITTI:}
One of the most popular datasets for autonomous driving, KITTI contains
7,481 training/validation images, and 7,518 for testing with held
annotations. The 2D object detection task contains three categories: car,
pedestrian, and cyclist. Evaluation is based on the VOC AP at IoU of 0.7,
 0.5, and 0.5 for the three categories, respectively. Since the focus
of this work is high quality detection, the Cascade R-CNN was only
tested on the car category. As shown in Table \ref{tab:kitti},
it improved the baseline by 0.87 points for the \texttt{Moderate}, and 1.9
points for the \texttt{Hard} regime, on the test set. These improvements
are nontrivial, given that MS-CNN is a strong detector and the
KITTI \texttt{car} detection task is fairly saturated.

\vspace{0.2cm}
\noindent{\bf CityPersons:}
CityPersons is a recently published pedestrian detection dataset,
collected across multiple European cities. It contains 2,975 training and
500 validation images, and 1,575 images for testing with held annotations.
Evaluation is based on miss-rate (MR) at IoU=0.5. We also report results
for MR at IoU=0.75, which is more commensurate with high quality detection.
This is consistent with a recent trend to adopt the stricter COCO metric
for pedestrian and face detection, see e.g. the Wider Challenge
2018. Table \ref{tab:citypersons} compares the validation set performance of the
Cascade R-CNN with that of the baseline MS-CNN (performances on validation
and test sets are usually equivalent on this dataset). The Cascade R-CNN
has large performance gains, especially for the stricter evaluation metric.
For example, it improves the baseline performance by $\sim$10 points
on the \texttt{Reasonable} set at MR$_{75}$.

\begin{table}[t]
\tablestyle{1.8pt}{1.2}
\begin{tabular}{c|c|c|x{22}x{22}x{22}}
&backbone &cascade &AP & AP$_{50}$ & AP$_{75}$\\ [.1em]
\shline
\multirow{2}{*}{Faster R-CNN} &\multirow{2}{*}{AlexNet} & \xmark &29.4 &63.2  &23.7\\
& & \cmark &38.9 &66.5  &40.5\\\hline
\multirow{2}{*}{Faster R-CNN} &\multirow{2}{*}{VGG} & \xmark &42.9 &76.4  &44.1\\
& & \cmark &51.2 &79.1  &56.3\\\hline
\multirow{2}{*}{R-FCN} &\multirow{2}{*}{RetNet-50} & \xmark &44.8 &77.5  &46.8\\
& & \cmark &51.8 &78.5  &57.1\\\hline
\multirow{2}{*}{R-FCN} &\multirow{2}{*}{ResNet-101} & \xmark &49.4 &79.8  &53.2\\
& & \cmark &54.2 &79.6  &59.2\\\hline
\end{tabular}\vspace{2mm}
\caption{Detection results on PASCAL VOC 2007 \texttt{test}.}
\label{tab:voc}\vspace{-3mm}
\end{table}

\begin{table}[t]
\tablestyle{1.8pt}{1.2}
\begin{tabular}{x{22}|x{32}|x{22}x{36}x{22}}
&cascade &Easy &Moderate &Hard\\ [.1em]
\shline
\multirow{2}{*}{AP$_{70}$} & \xmark &90.22 &89.08  &76.50\\
& \cmark &90.68 &89.95  &78.40\\\hline
\end{tabular}\vspace{2mm}
\caption{MS-CNN detection results for the car class on KITTI test set.}
\label{tab:kitti}\vspace{-3mm}
\end{table}

\begin{table}[t]
\tablestyle{1.8pt}{1.2}
\begin{tabular}{x{22}|x{32}|x{38}x{28}x{28}x{28}}
&cascade &Reasonable &Small &Heavy &All\\ [.1em]
\shline
\multirow{2}{*}{MR$_{50}$} & \xmark &13.07 &40.42  &53.55 &38.74\\
&\cmark &11.96 &38.37  &49.41 &36.83\\\hline
\multirow{2}{*}{MR$_{75}$} & \xmark &38.23 &59.84  &85.06 &65.56\\
&\cmark &28.45 &56.24  &81.86 &58.24\\\hline
\end{tabular}\vspace{2mm}
\caption{MS-CNN detection results on CityPersons validation set.}
\label{tab:citypersons}\vspace{-3mm}
\end{table}

\vspace{0.2cm}
\noindent{\bf WiderFace:}
One of the most challenging face detection datasets, mainly due to its
diversity in scale, pose and occlusion, WiderFace contains 32,203 images
with 393,703 annotated faces, of which 12,880 are used for training,
3,226 for validation, and the remainder for testing with held annotations.
Evaluation is based on the VOC AP at IoU=0.5 on three subsets,
\texttt{easy}, \texttt{medium} and \texttt{hard}, of different detection
difficulty. Again, we have used AP at IoU=0.5 and IoU=0.75 and evaluation
on the validation set. Table \ref{tab:widerface} shows that, while the Cascade
R-CNN is close to the baseline MS-CNN for AP$_{50}$, it significantly
boosts its performance for AP$_{75}$. The gain is smaller
on the \texttt{hard} than on the \texttt{easy} and \texttt{medium},
because the former contains mainly very small and heavily occluded faces,
for which high quality detection is difficult. This observation
mirrors the COCO experiments of Table \ref{tab:generalization}, where
improvements in AP$_S$ are smaller than for AP$_L$.

\begin{table}[t]
\tablestyle{1.8pt}{1.2}
\begin{tabular}{x{22}|x{32}|x{22}x{36}x{22}}
&cascade &Easy &Medium &Hard\\ [.1em]
\shline
\multirow{2}{*}{AP$_{50}$} & \xmark &91.1 &90.6  &81.0\\
& \cmark &91.3 &90.3  &81.1\\\hline
\multirow{2}{*}{AP$_{75}$} & \xmark &59.7 &61.3  &40.7\\
& \cmark &68.7 &66.3  &42.8\\\hline
\end{tabular}\vspace{2mm}
\caption{MS-CNN Detection results on WiderFace validation set.}
\label{tab:widerface}\vspace{-3mm}
\end{table}

\section{Conclusion}

In this work, we have proposed a multi-stage object detection framework,
the Cascade R-CNN, for high quality object detection, a rarely explored
problem in the detection literature. This architecture was shown to overcome
the high quality detection challenges of overfitting during training and
quality mismatch during inference. This is achieved by
training stages sequentially, using the output of one to train the next, and the same cascade is applied at inference.
The Cascade R-CNN was shown to achieve very consistent performance gains
on multiple challenging datasets, including
COCO, PASCAL VOC, KITTI, CityPersons, and WiderFace, for both generic
and specific object detection. These gains were also observed for
many object detectors, backbone networks, and techniques for
detection and instance segmentation. We thus believe that the Cascade
R-CNN can be useful for many future object detection and instance
segmentation research efforts.

\vspace{0.2cm}
\noindent{\bf Acknowledgment:}
This work was funded by NSF Awards IIS-1546305 and IIS-1637941, and a
GPU donation from NVIDIA. We would also
like to thank Kaiming He for valuable discussions.

{\small
\bibliographystyle{ieee}
\bibliography{egbib}
}


%



\ifCLASSOPTIONcaptionsoff
  \newpage
\fi

%
%

%





\end{document}